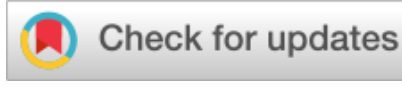

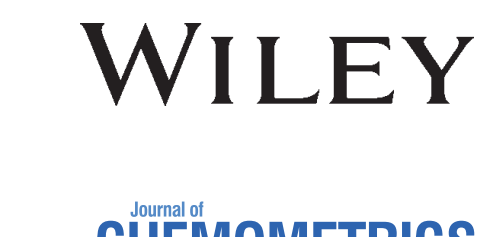

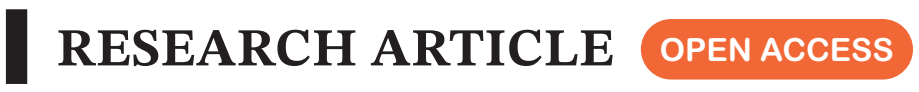

# Fast Partition-Based Cross-Validation With Centering and Scaling for $\mathbf{X}^T\mathbf{X}$ and $\mathbf{X}^T\mathbf{Y}$

Ole-Christian Galbo Engstrøm[1,2,3] | Martin Holm Jensen[1]

[1]Research and Development, FOSS Analytical A/S, Hillerød, Denmark | [2]Department of Computer Science, University of Copenhagen, Copenhagen, Denmark | [3]Department of Food Science, University of Copenhagen, Frederiksberg, Denmark

**Correspondence:** Ole-Christian Galbo Engstrøm (ole.e@di.ku.dk)

**Received:** 11 September 2024 | **Revised:** 27 January 2025 | **Accepted:** 28 January 2025

**Funding:** This research was supported by the Innovation Fund Denmark and FOSS Analytical A/S, grant number: 1044-00108B.

## ABSTRACT

We present algorithms that substantially accelerate partition-based cross-validation for machine learning models that require matrix products $\mathbf{X}^T\mathbf{X}$ and $\mathbf{X}^T\mathbf{Y}$. Our algorithms have applications in model selection for, for example, principal component analysis (PCA), principal component regression (PCR), ridge regression (RR), ordinary least squares (OLS), and partial least squares (PLS). Our algorithms support all combinations of column-wise centering and scaling of $\mathbf{X}$ and $\mathbf{Y}$, and we demonstrate in our accompanying implementation that this adds only a manageable, practical constant over efficient variants without preprocessing. We prove the correctness of our algorithms under a fold-based partitioning scheme and show that the running time is independent of the number of folds; that is, they have the same time complexity as that of computing $\mathbf{X}^T\mathbf{X}$ and $\mathbf{X}^T\mathbf{Y}$ and space complexity equivalent to storing $\mathbf{X}$, $\mathbf{Y}$, $\mathbf{X}^T\mathbf{X}$, and $\mathbf{X}^T\mathbf{Y}$. Importantly, unlike alternatives found in the literature, we avoid data leakage due to preprocessing. We achieve these results by eliminating redundant computations in the overlap between training partitions. Concretely, we show how to manipulate $\mathbf{X}^T\mathbf{X}$ and $\mathbf{X}^T\mathbf{Y}$ using only samples from the validation partition to obtain the preprocessed training partition-wise $\mathbf{X}^T\mathbf{X}$ and $\mathbf{X}^T\mathbf{Y}$. To our knowledge, we are the first to derive correct and efficient cross-validation algorithms for any of the 16 combinations of column-wise centering and scaling, for which we also prove only 12 give distinct matrix products.

## 1 | Introduction

In machine learning and predictive modeling, the objective is often to discover a latent representation, relationship (model), or both for a set of data represented by data set matrices $\mathbf{X}$ and $\mathbf{Y}$. Both matrices have $N$ rows, where each row represents a (data) sample. The $\mathbf{X}$ matrix contains $K$ features (variables) in its columns, while the $\mathbf{Y}$ matrix contains $M$ observations (responses) in its columns. Many popular models use either or both of the matrix products[1] $\mathbf{X}^T\mathbf{X}$ and $\mathbf{X}^T\mathbf{Y}$ with examples including principal component analysis (PCA) [4], principal component regression (PCR) [5, 6], ridge regression (RR) [7], ordinary least squares (OLS) via normal equations [8], and various partial least squares regression (PLS-R) [9, 10], and partial least squares discriminant analysis (PLS-DA) [11–13] algorithms.

Cross-validation [14, 15] assists with model selection for such methods by robustly evaluating performance on unseen data and avoiding overfitting hyperparameter selection. The (sample-wise) $P$-fold cross-validation scheme splits samples into $P$ nonoverlapping validation partitions, each associated with a training partition comprising the samples not in the validation partition (i.e., the remaining $P-1$ validation partitions). This necessitates computing matrix products over the submatrices formed by the training partition of each fold of the data set and evaluating on the validation partition.

Therefore, a baseline (naive) algorithm increases the running time of model selection by a factor of $P$. Lindgren et al. [1], with improvements by Lindgren et al. [16], remedy this by computing matrix products once over the entire data set and storing submatrices of $\mathbf{X}^\mathbf{T}\mathbf{X}$ and $\mathbf{X}^\mathbf{T}\mathbf{Y}$ for each validation partition, which are subtracted from the data set matrix products to obtain the training partition matrix products. This faster algorithm has time complexity independent of $P$. It is not difficult to adapt the approach by Lindgren et al. [16] to avoid storing the $P$ submatrices (with dimensions $K \times K$ and $K \times M$) as demonstrated, for example, by Gianola and Schön [17], thereby reducing space complexity.

Often, we want to center by subtracting column-wise means and scale by dividing with the column-wise sample standard deviations so that each column in $\mathbf{X}$ and $\mathbf{Y}$ has a mean of 0 and variance of 1 (or slightly smaller with Bessel's correction). For cross-validation, these statistics should be computed over the training partition to avoid the risk of leakage by preprocessing, a common pitfall in machine learning which may overestimate performance and lead to issues with reproducibility [18, 19]. However, this is not the case for the preprocessing done in the fast algorithms by Lindgren et al. [16] or when $\mathbf{X}$ and $\mathbf{Y}$ are simply centered and scaled over the entire data set initially. Generally, such approaches yield training partition matrix products different from the expensive baseline algorithm, which we illustrate at the end of Section 4.1.

In this paper, we develop a fast algorithm (Algorithm 7) where centering and scaling are performed over training partition submatrices, enabling efficient model selection using $P$-fold cross-validation that is not compromised by information from outside the training partition. Correctness is shown in Proposition 17. Algorithm 7 is asymptotically independent of $P$, having time complexity $\Theta(NK(K+M))$ (Proposition 19, matching Lindgren et al. [16]) and space complexity $\Theta((K+N)(K+M))$ (Proposition 22, matching Gianola and Schön [17]). Note that the time bound is the same as that of computing $\mathbf{X}^\mathbf{T}\mathbf{X}$ and $\mathbf{X}^\mathbf{T}\mathbf{Y}$, and the space bound is the same as storing $\mathbf{X}$, $\mathbf{Y}$, $\mathbf{X}^\mathbf{T}\mathbf{X}$, and $\mathbf{X}^\mathbf{T}\mathbf{Y}$. This suggests that our algorithms inexpensively enable $P$-fold cross-validation with centering and scaling, which is validated empirically in Section 7.

To compute training partition statistics efficiently, we determine how to void the contribution from the validation partition on the statistics over the entire data set. The spirit of this approach is, therefore, the same as the fractionated subparts of Lindgren et al. [16] and akin to what Liland et al. [20] enable for centering of $\mathbf{X}\mathbf{X}^\mathbf{T}$ (useful for wide data sets). To our knowledge, the algorithms we develop are the first to achieve fast $P$-fold cross-validation while properly centering, scaling, or both, using statistics exclusively over the training partition. This finds application in model selection for, for example, PCA, PCR, RR, OLS, and partial least squares (PLS), especially for tall data sets ($N$ greater than $K$) where computing matrix products may well dominate the time required to fit models.

Section 2 discusses related work. We introduce fast algorithms for cross-validation incrementally in Section 3 (no preprocessing), Section 4 (column-wise centering), and Section 5 (column-wise centering and scaling), showing correctness and computational complexity. Our results apply generally to any preprocessing combination of centering and scaling, which we discuss in Section 6. Interestingly, whether centering $\mathbf{X}$, $\mathbf{Y}$, or both, each combination results in the same matrix product $\mathbf{X}^\mathbf{T}\mathbf{Y}$ (Lemma 9). We present results of benchmarking baseline and fast algorithms for all combinations of preprocessing in Section 7 (implementation by Engstrøm [21] under the permissive Apache 2.0 license) before concluding in Section 8.

## 2 | Related Work

Fast $P$-fold cross-validation for methods that require computing $\mathbf{X}^\mathbf{T}\mathbf{X}$ and $\mathbf{X}^\mathbf{T}\mathbf{Y}$ has been studied over past decades [1, 16, 17, 22], where efficiency is achieved by computing matrix products only over the validation partition for each fold. This yields an asymptotic running time of $\Theta(NK(K+M))$ matching the cost of computing $\mathbf{X}^\mathbf{T}\mathbf{X}$ and $\mathbf{X}^\mathbf{T}\mathbf{Y}$ and so is independent of the choice of $P$. In these methods, centering and scaling are based on statistics not computed over the particular training partition of the fold, thereby risking leakage from preprocessing. Naively extending these algorithms by computing statistics over the training partition submatrices increases the running time by a factor of $P$, taking it into baseline algorithm territory.

Combining cross-validation with models relying on either $\mathbf{X}^\mathbf{T}\mathbf{X}$, $\mathbf{X}^\mathbf{T}\mathbf{Y}$, or both has been studied for PCA [23–29], PCR [28, 30], RR [31], OLS [28], and PLS [26, 28, 30, 32]. These are cases where our results apply. Our algorithms are also readily applicable to speed up the inner loop in repeated double cross-validation [33]. A library for fast cross-validation with centering and scaling for the fast [34] and numerically stable [35] improved kernel PLS algorithms by Dayal and MacGregor [36] that has been made available by Engstrøm et al. [37], which shows practical superiority in terms of execution time over the baseline approach. This paper is the accompanying theoretical treatise of the efficient training partition-wise matrix products implemented by Engstrøm et al. [37].

Cross-validation is similarly useful for models that use the matrix product $\mathbf{X}\mathbf{X}^\mathbf{T}$ [20, 38, 39], specifically for wide data sets. Liland et al. [20] give a fast solution that supports training partition centering of $\mathbf{X}\mathbf{X}^\mathbf{T}$, though the case of scaling is not handled. When $N < K$, $\mathbf{X}\mathbf{X}^\mathbf{T}$ is smaller than $\mathbf{X}^\mathbf{T}\mathbf{X}$, and so practitioners would likely prefer computing $\mathbf{X}\mathbf{X}^\mathbf{T}$ which our algorithms do not extend to.

Related also is *column-wise* cross-validation, which considers subsets of columns of $\mathbf{X}$ and can be used for feature selection, a fast version of which is given by Stefansson et al. [40]. There are no additional concerns with centering and scaling for column-wise cross-validation, as statistics are computed independently for each column. The column-wise cross-validation approach can be used separately or carried out for each fold in sample-wise $P$-fold cross-validation, in which case our results directly apply.

Another set of preprocessing methods [41] that operate on each sample individually include standard normal variate (SNV) [42], detrend [42], and convolution with, for example, Savitzky–Golay (SG) filters [43, 44]. If such sample-wise preprocessing is desired, it must be applied before centering and scaling, and in this case, it is fully compatible with our results. This is unlike multiplicative scatter correction (MSC) [45] since this preprocessing method modifies samples based on a fitting over the training partition, an immediate version of which degrades the time complexity of our fast algorithms.

## 3 | Cross-Validation Without Preprocessing

We proceed incrementally towards the fast cross-validation algorithm supporting training partition centering and scaling in Section 5 by first considering the case without preprocessing in this section and in Section 4 the case with centering.

An entry $(i,j)$ in a matrix $\mathbf{M}$ is denoted $\mathbf{M}_{ij}$ and is the entry in the $i$'th row and $j$'th column of $\mathbf{M}$. Let $R = \{1, \ldots, N\}$ be the set of indices (rows) and $S \subseteq R$, then we denote by $\mathbf{M}_S$ the submatrix of $\mathbf{M}$ containing each index (row) in $S$. We omit parentheses and write $\mathbf{M}_S^{\mathbf{T}}$ to mean $(\mathbf{M}_S)^{\mathbf{T}}$, $\mathbf{M}_{S_{ij}}$ to mean $(\mathbf{M}_S)_{ij}$, and $\mathbf{M}_{S_{ij}}^{\mathbf{T}}$ to mean $\left((\mathbf{M}_S)^{\mathbf{T}}\right)_{ij}$. To refer to the $i$'th row of a matrix, we write $\mathbf{M}_{i*}$, and for the $j$'th column, we write $\mathbf{M}_{*j}$. For visual clarity, we may write $\mathbf{M}_i$ instead of $\mathbf{M}_{i*}$ and emphasize that in such cases, we're always indexing *rows* of the matrix. Data set matrices are $\mathbf{X} \in \mathbb{R}^{N \times K}$ with $N$ rows and $K$ features, and $\mathbf{Y} \in \mathbb{R}^{N \times M}$ with $N$ rows and $M$ observations. We refer to $\mathbf{X}$, $\mathbf{Y}$, the set of rows $R$, and dimensions $N, K, M$ as given in this paragraph in the remainder.

**Definition 1.** training partition, validation partition Let $2 \leq P \leq N$ be the number of (cross-validation) partitions (folds). $P$-fold cross-validation associates each row $n \in R$ with a *partition* $p \in \{1, \ldots, P\}$, indicating which partition $n$ belongs to. A (valid) *partitioning* $\mathcal{P}$ of $R$ is an array of length $N$ such that $\bigcup_{n \in R} \mathcal{P}[n] = \{1, \ldots, P\}$ and $\mathcal{P}[n] = p$ means $n$ belongs to $p$. For $p \in \{1, \ldots, P\}$ and *partitioning* $\mathcal{P}$, the validation partition $V_p$ is given by $\{n \in R \mid \mathcal{P}[n] = p\}$, and the training partition $T_p$ by $R \setminus V_p$.

In the remainder, we will refer to $P$ as the number of partitions (folds). Observe that from Definition 1 follows $T_p \cap V_p = \emptyset$, $T_p \cup V_p = R$, and $\sum_{p=1}^{P} |V_p| = N$ since all $V_p$ are pairwise disjoint. These properties describe $P$-fold cross-validation, where the case of $P = N$ is leave-one-out cross-validation. Definition 1 also admits $\mathcal{P}[n] = \left((n-1) \bmod P\right) + 1$ that places every $P$'th sample in the same partition as in venetian blinds cross-validation and $\mathcal{P}[n] = \left\lfloor \frac{n-1}{P} \right\rfloor + 1$ that describes $P$-block-sized contiguous block cross-validation. Observe that we do not assume validation partitions are balanced; that is, they do not need to contain roughly the same amount of samples. Algorithm 1 is used for preprocessing a partitioning and store each $V_p$.

We consider Algorithm 2 the baseline method for computing matrix products $\mathbf{X}_T^{\mathbf{T}}\mathbf{X}_T$ and $\mathbf{X}_T^{\mathbf{T}}\mathbf{Y}_T$ for each training partition $T$, with the more efficient alternative being Algorithm 3 using the fractionated subparts insight of Lindgren et al. [16]. To clarify our analysis, we store intermediate computations in variables and allow for the same notation on these variables as we do for matrices. For instance, when $\mathrm{X}_T$ is assigned $\mathbf{X}_T$ and we write $\mathrm{XTX}_T \leftarrow \mathrm{X}_T^{\mathbf{T}}\mathrm{X}_T$, then $\mathrm{X}_T^{\mathbf{T}}$ is the transposition of $\mathrm{X}_T$ (equal to $\mathbf{X}_T^{\mathbf{T}}$), and $\mathrm{XTX}_T$ is effectively assigned the value $\mathbf{X}_T^{\mathbf{T}}\mathbf{X}_T$. Below, we show the correctness of the algorithms and analyze their computational complexities.

### 3.1 | Correctness

**Lemma 1.** *Let $\mathcal{P}$ be a partitioning of $R$ for $P$ partitions and $\mathcal{V}$ the array computed by Algorithm 1. Then for any $n \in R$ and $p \in \{1, \ldots, P\}$, $n \in \mathcal{V}[p]$ iff $\mathcal{P}[n] = p$.*

*Proof.* Consider any $p \in \{1, \ldots, P\}$. Step 1 ensures that $\mathcal{V}[p]$ is initially the empty set. Steps 3 and 4 extend $\mathcal{V}[p]$ with $n$ iff $\mathcal{P}[n] = p$. From step 2, we consider every $n \in R$, so this holds for all $n \in R$ and $p \in \{1, \ldots, P\}$. □

**Algorithm 1** Compute Validation Partitions

**Input:** $\mathcal{P}$ is a valid partitioning of $R = \{1, \ldots, N\}$

1: Let $\mathcal{V}$ be an array of length $P$ where each element is the empty set
2: **for** $n \in R$ **do**
3: $\quad p \leftarrow \mathcal{P}[n]$ ▷ Obtain $p$ which $n$ belongs to
4: $\quad \mathcal{V}[p] \leftarrow \mathcal{V}[p] \cup \{n\}$ ▷ Add $n$ to $V_p$
5: **end for**
6: **return** $\mathcal{V}$

**Algorithm 2** Baseline Cross-Validation Algorithm

**Input:** $\mathbf{X} \in \mathbb{R}^{N \times K}$, $\mathbf{Y} \in \mathbb{R}^{N \times M}$, $\mathcal{P}$ is a valid partitioning of $R = \{1, \ldots, N\}$

1: $R \leftarrow \{1, \ldots, N\}$
2: Let $\mathcal{V}$ be the result of executing Algorithm 1 with $\mathcal{P}$ and $R$ as input.
3: **for** $p \in \{1, \ldots, P\}$ **do**
4: $\quad V \leftarrow \mathcal{V}[p]$
5: $\quad T \leftarrow R \setminus V$
6: $\quad \mathrm{X}_T \leftarrow \mathbf{X}_T$, $\mathrm{Y}_T \leftarrow \mathbf{Y}_T$
7: $\quad \mathrm{XTX}_T \leftarrow \mathrm{X}_T^{\mathbf{T}}\mathrm{X}_T$, $\mathrm{XTY}_T \leftarrow \mathrm{X}_T^{\mathbf{T}}\mathrm{Y}_T$ ▷ Obtain $\mathbf{X}_T^{\mathbf{T}}\mathbf{X}_T$ and $\mathbf{X}_T^{\mathbf{T}}\mathbf{Y}_T$
8: **end for**

**Algorithm 3** Fast Cross-Validation Algorithm

**Input:** $\mathbf{X} \in \mathbb{R}^{N \times K}$, $\mathbf{Y} \in \mathbb{R}^{N \times M}$, $\mathcal{P}$ is a valid partitioning of $R = \{1, \ldots, N\}$

1: $R \leftarrow \{1, \ldots, N\}$
2: Let $\mathcal{V}$ be the result of executing Algorithm 1 with $\mathcal{P}$ and $R$ as input.
3: $\text{XTX} \leftarrow \mathbf{X}^{\mathbf{T}}\mathbf{X}$, $\text{XTY} \leftarrow \mathbf{X}^{\mathbf{T}}\mathbf{Y}$
4: **for** each partition $p \in \{1, \ldots, P\}$ **do**
5: $\quad V \leftarrow \mathcal{V}[p]$
6: $\quad \text{X}_V \leftarrow \mathbf{X}_V$, $\text{Y}_V \leftarrow \mathbf{Y}_V$
7: $\quad \text{XTX}_V \leftarrow \text{X}_V^{\text{T}}\text{X}_V$, $\text{XTY}_V \leftarrow \text{X}_V^{\text{T}}\text{Y}_V$ $\qquad \triangleright$ Obtain $\mathbf{X}_V^{\mathbf{T}}\mathbf{X}_V$ and $\mathbf{X}_V^{\mathbf{T}}\mathbf{Y}_V$
8: $\quad \text{XTX}_T \leftarrow \text{XTX} - \text{XTX}_V$, $\text{XTY}_T \leftarrow \text{XTY} - \text{XTY}_V$ $\qquad \triangleright$ Obtain $\mathbf{X}_T^{\mathbf{T}}\mathbf{X}_T$ and $\mathbf{X}_T^{\mathbf{T}}\mathbf{Y}_T$
9: **end for**

Proposition 1 shows that the set $\mathcal{V}[p]$ produced by Algorithm 1 is the validation partition $V_p$ according to Definition 1. To show the correctness of Algorithm 3, we prove the fractionated subparts insight by [1], namely, that training partition matrix products can be obtained by subtracting validation partition matrix products from the data set matrix products.

**Lemma 2.** *Given $p \in \{1, \ldots, P\}$, let $T = T_p$ be a training partition and $V = V_p$ validation partition, then $\mathbf{X}_T^{\mathbf{T}}\mathbf{X}_T = \mathbf{X}^{\mathbf{T}}\mathbf{X} - \mathbf{X}_V^{\mathbf{T}}\mathbf{X}_V$.*

*Proof.* We show the equivalent statement $\mathbf{X}_T^{\mathbf{T}}\mathbf{X}_T + \mathbf{X}_V^{\mathbf{T}}\mathbf{X}_V = \mathbf{X}^{\mathbf{T}}\mathbf{X}$. Recall that each of $\mathbf{X}^{\mathbf{T}}\mathbf{X}$, $\mathbf{X}_T^{\mathbf{T}}\mathbf{X}_T$ and $\mathbf{X}_V^{\mathbf{T}}\mathbf{X}_V$ are $K \times K$ matrices and consider any entry $(i, j)$ in $\mathbf{X}^{\mathbf{T}}\mathbf{X}$ for $i \in \{1, \ldots, K\}$ and $j \in \{1, \ldots, K\}$ as given by,

$$\left(\mathbf{X}^{\mathbf{T}}\mathbf{X}\right)_{ij} = \sum_{n \in R} \mathbf{X}_{in}^{\mathbf{T}}\mathbf{X}_{nj}.$$

Similarly, we have

$$\left(\mathbf{X}_T^{\mathbf{T}}\mathbf{X}_T\right)_{ij} = \sum_{n \in T} \mathbf{X}_{in}^{\mathbf{T}}\mathbf{X}_{nj} \text{ and } \left(\mathbf{X}_V^{\mathbf{T}}\mathbf{X}_V\right)_{ij} = \sum_{n \in V} \mathbf{X}_{in}^{\mathbf{T}}\mathbf{X}_{nj}.$$

By Definition 1, $T \cap V = \emptyset$ and $T \cup V = R$, so with commutativity of addition, we have

$$\begin{aligned}\left(\mathbf{X}_T^{\mathbf{T}}\mathbf{X}_T\right)_{ij} + \left(\mathbf{X}_V^{\mathbf{T}}\mathbf{X}_V\right)_{ij} &= \sum_{n \in T} \mathbf{X}_{in}^{\mathbf{T}}\mathbf{X}_{nj} + \sum_{n \in V} \mathbf{X}_{in}^{\mathbf{T}}\mathbf{X}_{nj} \\ &= \sum_{n \in T \cup V} \mathbf{X}_{in}^{\mathbf{T}}\mathbf{X}_{nj} = \sum_{n \in R} \mathbf{X}_{in}^{\mathbf{T}}\mathbf{X}_{nj} \\ &= \left(\mathbf{X}^{\mathbf{T}}\mathbf{X}\right)_{ij}.\end{aligned}$$

□

**Lemma 3.** *Given $p \in \{1, \ldots, P\}$, let $T = T_p$ be a training partition and $V = V_p$ validation partition, then $\mathbf{X}_T^{\mathbf{T}}\mathbf{Y}_T = \mathbf{X}^{\mathbf{T}}\mathbf{Y} - \mathbf{X}_V^{\mathbf{T}}\mathbf{Y}_V$.*

*Proof.* As proof of Lemma 2 with $j \in \{1, \ldots, M\}$. □

**Proposition 1.** *Correctness, Algorithm 3 In step 7 of Algorithm 2, $\text{XTX}_T = \mathbf{X}_T^{\mathbf{T}}\mathbf{X}_T$ and $\text{XTY}_T = \mathbf{X}_T^{\mathbf{T}}\mathbf{Y}_T$ and are identical to $\text{XTX}_T$ and $\text{XTY}_T$ computed in step 8 of Algorithm 3.*

*Proof.* By Lemma 1, both algorithms select validation partition $V$ according to Definition 1. For Algorithm 2, $T = R \setminus V$ by step 5, hence, clearly, $\text{XTX}_T = \mathbf{X}_T^{\mathbf{T}}\mathbf{X}_T$ and $\text{XTY}_T = \mathbf{X}_T^{\mathbf{T}}\mathbf{Y}_T$ in step 7. Since step 8 of Algorithm 3 effectively computes $\mathbf{X}^{\mathbf{T}}\mathbf{X} - \mathbf{X}_V^{\mathbf{T}}\mathbf{X}_V$ and $\mathbf{X}^{\mathbf{T}}\mathbf{Y} - \mathbf{X}_V^{\mathbf{T}}\mathbf{Y}_V$, it follows from Lemmas 2 and 3 that $\text{XTX}_T$ and $\text{XTY}_T$ computed in step 8 are identical to those computed in step 7 of Algorithm 2. □

## 3.2 | Computational Complexity

We analyze the asymptotic time and space complexity of Algorithms 2 and 3, showing the latter shaves a $\Theta(P)$ factor off of the running time with no increase in space complexity. To this end, we assume that matrix multiplication is done using the iterative algorithm that requires $ABC$ operations to multiply two matrices with dimensions $A \times B$ and $B \times C$. While there are algorithms with improved bounds (e.g., by Le Gall [46]), the iterative algorithm is used in practice as it enables optimizations (e.g., cache-friendly memory layout, loop unrolling, efficient use of single instruction, multiple data [SIMD]) that significantly improve hardware performance. We also disregard that by the symmetry of $\mathbf{X}^{\mathbf{T}}\mathbf{X}$, we could roughly halve the number of operations needed by computing and mirroring the upper or lower triangular matrix, as this constant does not impact $\Theta$ bounds.

**Lemma 4.** *Algorithm 1 requires $\Theta(N)$ operations.*

*Proof.* Storing inputs $\mathcal{P}$ and $R$ require $2N$ operations. Step 1 requires $P = O(N)$ (since $P \leq N$) operations to construct the array $\mathcal{V}$. Indexing into $\mathcal{P}$ (step 3) and $\mathcal{V}$ (step 4) requires $\Theta(1)$ operations. By using an efficient set implementation, $n$ can be added to $\mathcal{V}[p]$ with $\Theta(1)$ operations. We do so $N$ times by step 2; therefore, Algorithm 1 requires $2N + O(N) + \sum_{n \in R} \Theta(1) = \Theta(N)$ operations. □

**Lemma 5.** *Algorithm 1 requires storing $\Theta(N)$ entries.*

*Proof.* Storing inputs $\mathcal{P}$ and $R$ require $2N$ entries. Step 1 requires $P$ entries storing $\mathcal{V}$. Steps 3 and 4 temporarily store $O(1)$ entries to extend $\mathcal{V}$ with one additional entry. $\mathcal{V}$ is extended $N$ times due to step 2. Therefore, the algorithm stores $2N + O(1) + P + N = \Theta(N)$ entries since $P \leq N$. □

**Lemma 6.** *Given a partitioning $\mathcal{P}$ of $R$, $\sum_{p=1}^{P} |T_p| = N(P-1)$.*

*Proof.* By Definition 1, $|T_p| = N - |V_p|$ and $\sum_{p=1}^{P} |V_p| = N$; therefore,

$$\sum_{p=1}^{P} |T_p| = \sum_{p=1}^{P} N - |V_p| = \sum_{p=1}^{P} N - \sum_{p=1}^{P} |V_p| = PN - N = N(P-1).$$

□

**Proposition 2.** *Algorithm 2 requires $\Theta(PNK(K+M))$ operations.*

*Proof.* Storing inputs $\mathbf{X}, \mathbf{Y}$, and $\mathcal{P}$ requires $NK + NM + N = \Theta(N(K+M))$ operations. In step 1, we construct a set with $N$ elements using $N$ operations. In step 2, we compute $\mathcal{V}$ with $\Theta(N)$ operations by Lemma 4. Consider a partition $p \in \{1, \ldots, P\}$ with validation partition $V_p$ and training partition $T_p$. Step 4 uses $|V_p|$ operations for assignment, and step 5 uses $|V_p| + |T_p| = N$ operations to compute $T$. In step 6, we select $|T_p|$ rows from both $\mathbf{X}$ and $\mathbf{Y}$ using $|T_p|(K+M)$ operations. Step 7 requires $\Theta(|T_p|K(K+M))$ operations to compute $\mathbf{X}_{T_p}^{\mathbf{T}}\mathbf{X}_{T_p}$ and $\mathbf{X}_{T_p}^{\mathbf{T}}\mathbf{Y}_{T_p}$. Iterating over $\{1, \ldots, P\}$ in step 3, it follows from $\sum_{p=1}^{P} |V_p| = N$ and Lemma 6 that

$$\begin{aligned}
&\sum_{p=1}^{P} \Theta(|V_p|) + 2\Theta(N) + |T_p|(K+M) + \Theta(|T_p|K(K+M)) \\
&= \Theta(N) + \Theta(PN) + N(P-1)(K+M) + \Theta(PNK(K+M)) \\
&= \Theta(PN(K+M)) + \Theta(PNK(K+M)).
\end{aligned}$$

Thus, the total number of operations is $\Theta(PNK(K+M))$. □

**Proposition 3.** *Algorithm 3 requires $\Theta(NK(K+M))$ operations.*

*Proof.* Storing inputs $\mathbf{X}$, $\mathbf{Y}$, and $\mathcal{P}$ requires $\Theta(N(K+M))$ operations. Step 1 uses $N$ operations. Step 2 uses $\Theta(N)$ operations by Lemma 4. Step 3 computes $\mathbf{X}^{\mathbf{T}}\mathbf{X}$ and $\mathbf{X}^{\mathbf{T}}\mathbf{Y}$ requiring respectively $\Theta(KNK)$ and $\Theta(KNM)$ operations. So, up to and including step 3 requires a total of $\Theta(NK(K+M))$ operations.

Consider a partition $p \in \{1, \ldots, P\}$ with validation partition $V_p$ and training partition $T_p$. Step 5 uses $\Theta(|V_p|)$ operations for assignment. Step 6 selects $|V_p|$ rows from $\mathbf{X}$, respectively $\mathbf{Y}$, using $|V_p|(K+M)$ operations. Matrix multiplication in step 7 requires $\Theta(|V_p|K(K+M))$ operations. In step 8, we subtract matrices with dimensions $K \times K$ and $K \times M$ using $\Theta(K(K+M))$ operations. So steps 5–8 require $\Theta(|V_p|) + |V_p|(K+M) + \Theta(|V_p|K(K+M)) + \Theta(K(K+M)) = \Theta(|V_p|K(K+M))$ operations. Iterating over $\{1, \ldots, P\}$ in step 4, we have by $\sum_{p=1}^{P} |V_p| = N$ that

$$\begin{aligned}
\Theta\left(\sum_{p=1}^{P} |V_p|K(K+M)\right) &= \Theta\left(K(K+M) \cdot \sum_{p=1}^{P} |V_p|\right) \\
&= \Theta\big(NK(K+M)\big).
\end{aligned}$$

It follows that Algorithm 3 requires $\Theta(NK(K+M)) + \Theta(NK(K+M)) = \Theta(NK(K+M))$ operations. □

**Proposition 4.** *Algorithm 2 requires $\Theta(P)$ more operations than Algorithm 3.*

*Proof.* By Propositions 2 and 3, the ratio of operations is $\frac{\Theta(PNK(K+M))}{\Theta(NK(K+M))} = \Theta(P)$. □

Below, we show that the reduction in running time indicated by Proposition 4 does not come at the cost of increasing space usage.

**Proposition 5.** *Algorithm 2 requires storing $\Theta((K+N)(K+M))$ entries.*

*Proof.* Storing inputs $\mathbf{X}$, $\mathbf{Y}$, and $\mathcal{P}$ requires $\Theta(N(K+M))$ entries. Step 1 stores $N$ entries. Step 2 stores $\Theta(N)$ entries by Lemma 5. Steps 4-5 store $|V| + |T| = N$ entries. Step 6 stores $|T|K + |T|M$ entries, while step 7 requires $K^2 + KM$ entries. This totals $\Theta(N(K+M)) + N + \Theta(N) + N + |T|K + |T|M + K^2 + KM$ entries, which since $|T| < N$ is $\Theta(N(K+M)) + \Theta(K^2 + KM)$. Therefore, the number of entries stored at any step of Algorithm 2 is $\Theta((K+N)(K+M))$. □

**Proposition 6.** *Algorithm 3 requires storing of $\Theta((K+N)(K+M))$ entries.*

*Proof.* Storing inputs $\mathbf{X}$, $\mathbf{Y}$, and $\mathcal{P}$ requires $\Theta(N(K+M))$ entries. Step 1 stores $N$ entries. Step 2 stores $\Theta(N)$ entries by Lemma 5. For step 3, XTX and XTY require $K^2 + KM$ entries. This means $\Theta((K+N)(K+M)))$ prior to step 4. Step 5 requires $|V|$ entries. Step 6 requires $|V|K + |V|M$ entries. Steps 7 and 8 both take up $K^2 + KM$ entries. This totals $\Theta((K+N)(K+M))) + |V| + |V|K + |V|M + 2(K^2 + KM)$ entries. Since $|V| < N$, the number of entries stored at any step of Algorithm 3 is $\Theta((K+N)(K+M))$. □

## 4 | Cross-Validation With Centering

In many cases, column-wise centering of $\mathbf{X}$ and $\mathbf{Y}$ is employed as a preprocessing step to give individual features and observations a mean of zero. This not only simplifies the interpretation of model coefficients but also improves numerical stability, reduces bias from differing feature magnitudes, and may improve the convergence of optimization algorithms. While Algorithm 2 can be immediately extended to support this with no impact on asymptotic complexity, this is not so for Algorithm 3, since directly computing the mean over columns in $\mathbf{X}_T$ and $\mathbf{Y}_T$ reintroduces a dependency on $P$ as evidenced by Lemma 6.

While Lindgren et al. [16] propose a solution for centering that is more efficient in terms of running time, it falls short of capturing training partition centering as defined next. Indeed, their solution computes matrix products that do not match those of Algorithm 4 (baseline cross-validation algorithm with centering), as we explore further in Section 4.1. In contrast, Algorithm 5 performs training partition centering and shares asymptotic complexity with Algorithm 3. It works by computing the mean *once* for the entire data set and subtracting the contribution from samples in $V_p$ as we iterate over each partition $p$.

**Definition 2.** Let $S \subseteq R$ denote a set of rows. The mean row vector $\overline{\mathbf{x}_S}$ has width $K$ and is the row vector of column means of

**Algorithm 4** Baseline Cross-Validation Algorithm with Centering

**Input:** $\mathbf{X} \in \mathbb{R}^{N\times K}$, $\mathbf{Y} \in \mathbb{R}^{N\times M}$, $\mathcal{P}$ is a valid partitioning of $R = \{1, \dots, N\}$

1: $R \leftarrow \{1, \dots, N\}$

2: Let $\mathcal{V}$ be the result of executing Algorithm 1 with $\mathcal{P}$ and $R$ as input

3: **for** each partition $p \in \{1, \dots, P\}$ **do**

4: $\quad V \leftarrow \mathcal{V}[p]$

5: $\quad T \leftarrow R \setminus V$

6: $\quad \mathrm{X}_T \leftarrow \mathbf{X}_T,\ \mathrm{Y}_T \leftarrow \mathbf{Y}_T$

7: $\quad \mu\mathrm{X}_T \leftarrow \overline{\mathbf{X}_T},\ \mu\mathrm{Y}_T \leftarrow \overline{\mathbf{Y}_T}$

8: $\quad \mathrm{cX}_T \leftarrow \mathrm{X}_T - \mu\mathrm{X}_T,\ \mathrm{cY}_T \leftarrow \mathrm{Y}_T - \mu\mathrm{Y}_T$ $\quad\triangleright$ Obtain $\mathbf{X}_T^{\mathbf{c}}$ and $\mathbf{Y}_T^{\mathbf{c}}$

9: $\quad \mathrm{cXTX}_T \leftarrow \mathrm{cX}_T^{\mathbf{T}}\mathrm{cX}_T,\ \mathrm{cXTY}_T \leftarrow \mathrm{cX}_T^{\mathbf{T}}\mathrm{cY}_T$ $\quad\triangleright$ Obtain $\mathbf{X}_T^{\mathbf{cT}}\mathbf{X}_T^{\mathbf{c}}$ and $\mathbf{X}_T^{\mathbf{cT}}\mathbf{Y}_T^{\mathbf{c}}$

10: **end for**

**Algorithm 5** Fast Cross-Validation Algorithm with Centering

**Input:** $\mathbf{X} \in \mathbb{R}^{N\times K}$, $\mathbf{Y} \in \mathbb{R}^{N\times M}$, $\mathcal{P}$ is a valid partitioning of $R = \{1, \dots, N\}$

1: $R \leftarrow \{1, \dots, N\}$

2: Let $\mathcal{V}$ be the result of executing Algorithm 1 with $\mathcal{P}$ and $R$ as input

3: $\mathrm{XTX} \leftarrow \mathbf{X}^{\mathbf{T}}\mathbf{X},\ \mathrm{XTY} \leftarrow \mathbf{X}^{\mathbf{T}}\mathbf{Y},\ \mu\mathrm{x} \leftarrow \overline{\mathbf{x}},\ \mu\mathrm{y} \leftarrow \overline{\mathbf{y}}$

4: **for** each partition $p \in \{1, \dots, P\}$ **do**

5: $\quad V \leftarrow \mathcal{V}[p]$

6: $\quad \mathrm{X}_V \leftarrow \mathbf{X}_V,\ \mathrm{Y}_V \leftarrow \mathbf{Y}_V$

7: $\quad \mathrm{XTX}_V \leftarrow \mathrm{X}_V^{\mathbf{T}}\mathrm{X}_V,\ \mathrm{XTY}_V \leftarrow \mathrm{X}_V^{\mathbf{T}}\mathrm{Y}_V$

8: $\quad \mathrm{XTX}_T \leftarrow \mathrm{XTX} - \mathrm{XTX}_V,\ \mathrm{XTY}_T \leftarrow \mathrm{XTY} - \mathrm{XTY}_V$ $\quad\triangleright$ Obtain $\mathbf{X}_T^{\mathbf{T}}\mathbf{X}_T$ and $\mathbf{X}_T^{\mathbf{T}}\mathbf{Y}_T$

9: $\quad \mu\mathrm{x}_V \leftarrow \overline{\mathbf{x}_V},\ \mu\mathrm{y}_V \leftarrow \overline{\mathbf{y}_V}$

10: $\quad N_V \leftarrow |V|,\ N_T \leftarrow N - N_V$

11: $\quad \mu\mathrm{x}_T \leftarrow \frac{N}{N_T}\mu\mathrm{x} - \frac{N_V}{N_T}\mu\mathrm{x}_V$ $\quad\triangleright$ Obtain $\overline{\mathbf{x}_T}$

12: $\quad \mu\mathrm{y}_T \leftarrow \frac{N}{N_T}\mu\mathrm{y} - \frac{N_V}{N_T}\mu\mathrm{y}_V$ $\quad\triangleright$ Obtain $\overline{\mathbf{y}_T}$

13: $\quad \mu\mathrm{xTx}_T \leftarrow N_T\left(\mu\mathrm{x}_T^{\mathbf{T}}\mu\mathrm{x}_T\right),\ \mu\mathrm{xTy}_T \leftarrow N_T\left(\mu\mathrm{x}_T^{\mathbf{T}}\mu\mathrm{y}_T\right)$ $\quad\triangleright$ Obtain $|T|\overline{\mathbf{x}_T^{\mathbf{T}}}\overline{\mathbf{x}_T}$ and $|T|\overline{\mathbf{x}_T^{\mathbf{T}}}\overline{\mathbf{y}_T}$

14: $\quad \mathrm{cXTX}_T \leftarrow \mathrm{XTX}_T - \mu\mathrm{xTx}_T$ $\quad\triangleright$ Obtain $\mathbf{X}_T^{\mathbf{cT}}\mathbf{X}_T^{\mathbf{c}}$

15: $\quad \mathrm{cXTY}_T \leftarrow \mathrm{XTY}_T - \mu\mathrm{xTy}_T$ $\quad\triangleright$ Obtain $\mathbf{X}_T^{\mathbf{cT}}\mathbf{Y}_T^{\mathbf{c}}$

16: **end for**

$\mathbf{X}_S$, and similarly, $\overline{\mathbf{y}_S}$ of width $M$ is the mean row vector of $\mathbf{Y}_S$. We stack the mean row vectors to obtain matrices $\overline{\mathbf{X}_S}$ with dimension $|S| \times K$ and $\overline{\mathbf{Y}_S}$ with dimension $|S| \times M$, namely,

$$\overline{\mathbf{X}_S} = \begin{bmatrix} \overline{\mathbf{x}_S} \\ \vdots \\ \overline{\mathbf{x}_S} \end{bmatrix} \text{ and } \overline{\mathbf{Y}_S} = \begin{bmatrix} \overline{\mathbf{y}_S} \\ \vdots \\ \overline{\mathbf{y}_S} \end{bmatrix}.$$

For brevity, we adopt the convention that $\overline{\mathbf{X}_S^{\mathbf{T}}}$ and $\overline{\mathbf{x}_S^{\mathbf{T}}}$, respectively, are given by $\left(\overline{\mathbf{X}_S}\right)^{\mathbf{T}}$ and $\left(\overline{\mathbf{x}_S}\right)^{\mathbf{T}}$. We also omit parentheses to denote the $i$'th element of vector $\overline{\mathbf{x}_S^{\mathbf{T}}}$ by $\overline{\mathbf{x}_{S_i}^{\mathbf{T}}}$ and the $i$'th element of vector $\overline{\mathbf{y}_S}$ by $\overline{\mathbf{y}_{S_i}}$. We write $\overline{\mathbf{x}}$ to mean $\overline{\mathbf{x}_R}$ and $\overline{\mathbf{y}}$ to mean $\overline{\mathbf{y}_R}$ when clear from context.

**Definition 3.** Let $S \subseteq R$ denote a set of rows. The (column-wise) *centering* of $\mathbf{X}_S$ is $\mathbf{X}_S^{\mathbf{c}} = \mathbf{X}_S - \overline{\mathbf{X}_S}$, and the (column-wise) *centering* of $\mathbf{Y}_S$ is $\mathbf{Y}_S^{\mathbf{c}} = \mathbf{Y}_S - \overline{\mathbf{Y}_S}$.

To support column-wise centering of $\mathbf{X}_T$ and $\mathbf{Y}_T$ for a training partition $T$, we subtract the column-wise means before taking matrix products, namely,

$$\mathbf{X}_T^{\mathbf{cT}}\mathbf{X}_T^{\mathbf{c}} = \left(\mathbf{X}_T^{\mathbf{T}} - \overline{\mathbf{X}_T^{\mathbf{T}}}\right)\left(\mathbf{X}_T - \overline{\mathbf{X}_T}\right) \text{ and}$$
$$\mathbf{X}_T^{\mathbf{cT}}\mathbf{Y}_T^{\mathbf{c}} = \left(\mathbf{X}_T^{\mathbf{T}} - \overline{\mathbf{X}_T^{\mathbf{T}}}\right)\left(\mathbf{Y}_T - \overline{\mathbf{Y}_T}\right).$$

We refer to these quantities as centered matrix products, and they are the targets of Algorithm 4 and the more efficient Algorithm 5.

### 4.1 | Correctness

We proceed to show the correctness of Algorithm 5, where we note steps 1–8 are as in Algorithm 3 except also computing mean row vectors of $\mathbf{X}$ and $\mathbf{Y}$ in step 3. To illustrate the purpose of steps 9-15, consider a validation partition $V \subseteq R$ and training partition $T = R \setminus V$. We show in Proposition 7 that the centered matrix product $\mathbf{X}_T^{\mathbf{cT}}\mathbf{Y}_T^{\mathbf{c}}$ equals $\mathbf{X}_T^{\mathbf{T}}\mathbf{Y}_T - |T|\overline{\mathbf{x}_T^{\mathbf{T}}}\overline{\mathbf{y}_T}$. Just as Lemma 2 is used to efficiently compute the left-hand term with running time depending on $|V|$ rather than $|T|$, the same is possible for the right-hand term. Lemma 7 shows that $\overline{\mathbf{x}_T}$ equals $\frac{N}{|T|}\overline{\mathbf{x}} - \frac{|V|}{|T|}\overline{\mathbf{x}_V}$, so we need only compute $\overline{\mathbf{x}_V}$ for each partition (similarly for $\overline{\mathbf{y}_T}$).

**Lemma 7.** *Given a validation partition $V$ and a training partition $T = R \setminus V$, then $\overline{\mathbf{x}_T}$, the mean row vector of $\mathbf{X}_T$, is equal to $\frac{N}{|T|}\overline{\mathbf{x}_R} - \frac{|V|}{|T|}\overline{\mathbf{x}_V}$.*

*Proof.* Using Definition 2 for $\overline{\mathbf{x}_R}$ and $\overline{\mathbf{x}_V}$, we have

$$\begin{aligned} \frac{N}{|T|}\overline{\mathbf{x}_R} - \frac{|V|}{|T|}\overline{\mathbf{x}_V} &= \frac{N}{|T|}\frac{1}{N}\sum_{n\in R}\mathbf{x}_n - \frac{|V|}{|T|}\frac{1}{|V|}\sum_{n\in V}\mathbf{x}_n \\ &= \frac{1}{|T|}\sum_{n\in R}\mathbf{x}_n - \frac{1}{|T|}\sum_{n\in V}\mathbf{x}_n, \end{aligned}$$

and since $R = T \cup V$ and $T \cap V = \emptyset$, we can rearrange the sum over $R$,

$$\frac{1}{|T|}\sum_{n\in T}\mathbf{X}_n + \frac{1}{|T|}\sum_{n\in V}\mathbf{X}_n - \frac{1}{|T|}\sum_{n\in V}\mathbf{X}_n = \frac{1}{|T|}\sum_{n\in T}\mathbf{X}_n = \overline{\mathbf{x}_T}.$$

□

**Lemma 8.** *Given a validation partition $V$ and a training partition $T = R \setminus V$, then $\overline{\mathbf{y}_T}$, the mean row vector of $\mathbf{Y}_T$, is equal to $\frac{N}{|T|}\overline{\mathbf{y}_R} - \frac{|V|}{|T|}\overline{\mathbf{y}_V}$.*

*Proof.* As proof of Lemma 7. □

We note that Lemma 9 is interesting beyond the purpose of proving correctness, stating that whether you center either $\mathbf{X}$, $\mathbf{Y}$, or both, when centering is performed, the result is $\mathbf{X}_T^{\mathbf{cT}}\mathbf{Y}_T^{\mathbf{c}}$ (i.e., the same centered matrix product), which we explore further in Section 6.

**Lemma 9.** *Let $S \subseteq R$, then the matrix products $\overline{\mathbf{X}_S^{\mathbf{T}}}\,\overline{\mathbf{Y}_S}$, $\overline{\mathbf{X}_S^{\mathbf{T}}}\mathbf{Y}_S$ and $\mathbf{X}_S^{\mathbf{T}}\overline{\mathbf{Y}_S}$ are all equal to $|S|\overline{\mathbf{x}_S^{\mathbf{T}}}\,\overline{\mathbf{y}_S}$. That is, $\mathbf{X}_S^{\mathbf{T}}\overline{\mathbf{Y}_S} = \overline{\mathbf{X}_S^{\mathbf{T}}}\mathbf{Y}_S = \overline{\mathbf{X}_S^{\mathbf{T}}}\,\overline{\mathbf{Y}_S} = |S|\overline{\mathbf{x}_S^{\mathbf{T}}}\,\overline{\mathbf{y}_S}$.*

*Proof.* Let $(i,j)$ be an entry in $\overline{\mathbf{X}_S^{\mathbf{T}}}\,\overline{\mathbf{Y}_S}$. Observe that by Definition 2, each *column* in $\overline{\mathbf{X}_S^{\mathbf{T}}}$ is equal to the column vector $\overline{\mathbf{x}_S^{\mathbf{T}}}$; thus, for all $n \in S$, we have $\overline{\mathbf{x}_S^{\mathbf{T}}} = \overline{\mathbf{X}_S^{\mathbf{T}}}_{*n}$ and so $\overline{\mathbf{x}_S^{\mathbf{T}}}_i = \overline{\mathbf{X}_S^{\mathbf{T}}}_{in}$. Similarly, for all $n \in S$, each *row* in $\overline{\mathbf{Y}_S}$ is the mean row vector $\overline{\mathbf{y}_S} = \overline{\mathbf{Y}_S}_{n*}$ and so $\overline{\mathbf{y}_S}_j = \overline{\mathbf{Y}_S}_{nj}$.

**(a)** By Definition 2 and distributivity of multiplication over addition, it follows that

$$\begin{aligned}
\left(\overline{\mathbf{X}_S^{\mathbf{T}}}\,\overline{\mathbf{Y}_S}\right)_{ij} &= \sum_{n\in S}\overline{\mathbf{X}_S^{\mathbf{T}}}_{in}\overline{\mathbf{Y}_S}_{nj} = \sum_{n\in S}\overline{\mathbf{x}_S^{\mathbf{T}}}_i\overline{\mathbf{y}_S}_j \\
&= \overline{\mathbf{x}_S^{\mathbf{T}}}_i\overline{\mathbf{y}_S}_j\sum_{n\in S}1 = |S|\overline{\mathbf{x}_S^{\mathbf{T}}}_i\overline{\mathbf{y}_S}_j \\
&= |S|\left(\overline{\mathbf{x}_S^{\mathbf{T}}}\,\overline{\mathbf{y}_S}\right)_{ij},
\end{aligned}$$

concluding the proof that $\overline{\mathbf{X}_S^{\mathbf{T}}}\,\overline{\mathbf{Y}_S} = |S|\overline{\mathbf{x}_S^{\mathbf{T}}}\,\overline{\mathbf{y}_S}$.

**(b)** By Definition 2 we have $\overline{\mathbf{y}_S}_j = \overline{\mathbf{Y}_S}_{nj} = \frac{1}{|S|}\sum_{n\in S}\mathbf{Y}_{S_{nj}}$. So by distributivity of multiplication over addition, we get

$$\begin{aligned}
|S|\overline{\mathbf{x}_S^{\mathbf{T}}}_i\overline{\mathbf{y}_S}_j &= |S|\overline{\mathbf{x}_S^{\mathbf{T}}}_i\frac{1}{|S|}\sum_{n\in S}\mathbf{Y}_{S_{nj}} = \overline{\mathbf{x}_S^{\mathbf{T}}}_i\sum_{n\in S}\mathbf{Y}_{S_{nj}} \\
&= \sum_{n\in S}\overline{\mathbf{x}_S^{\mathbf{T}}}_i\mathbf{Y}_{S_{nj}} = \sum_{n\in S}\overline{\mathbf{X}_S^{\mathbf{T}}}_{in}\mathbf{Y}_{S_{nj}} \\
&= \left(\overline{\mathbf{X}_S^{\mathbf{T}}}\mathbf{Y}_S\right)_{ij},
\end{aligned}$$

showing that $|S|\overline{\mathbf{x}_S^{\mathbf{T}}}\,\overline{\mathbf{y}_S} = \overline{\mathbf{X}_S^{\mathbf{T}}}\mathbf{Y}_S$.

**(c)** By Definition 2, we have $\overline{\mathbf{x}_S^{\mathbf{T}}}_i = \overline{\mathbf{X}_S^{\mathbf{T}}}_{in} = \frac{1}{|S|}\sum_{n\in S}\mathbf{X}_{S_{in}}^{\mathbf{T}}$. Similar to the argument in **(b)** and by commutativity of scalar multiplication, it follows that

$$\begin{aligned}
|S|\overline{\mathbf{x}_S^{\mathbf{T}}}_i\overline{\mathbf{y}_S}_j &= |S|\overline{\mathbf{y}_S}_j\frac{1}{|S|}\sum_{n\in S}\mathbf{X}_{S_{in}}^{\mathbf{T}} = \overline{\mathbf{y}_S}_j\sum_{n\in S}\mathbf{X}_{S_{in}}^{\mathbf{T}} \\
&= \sum_{n\in S}\mathbf{X}_{S_{in}}^{\mathbf{T}}\overline{\mathbf{y}_S}_j = \sum_{n\in S}\mathbf{X}_{S_{in}}^{\mathbf{T}}\overline{\mathbf{Y}_S}_{nj} \\
&= \left(\mathbf{X}_S^{\mathbf{T}}\overline{\mathbf{Y}_S}\right)_{ij};
\end{aligned}$$

hence, $|S|\overline{\mathbf{x}_S^{\mathbf{T}}}\,\overline{\mathbf{y}_S} = \mathbf{X}_S^{\mathbf{T}}\overline{\mathbf{Y}_S}$. Combining **(a)**, **(b)**, and **(c)**, it follows that $\mathbf{X}_S^{\mathbf{T}}\overline{\mathbf{Y}_S} = \overline{\mathbf{X}_S^{\mathbf{T}}}\mathbf{Y}_S = \overline{\mathbf{X}_S^{\mathbf{T}}}\,\overline{\mathbf{Y}_S} = |S|\overline{\mathbf{x}_S^{\mathbf{T}}}\,\overline{\mathbf{y}_S}$.

□

**Lemma 10.** *Let $S \subseteq R$, then the matrix products $\overline{\mathbf{X}_S^{\mathbf{T}}}\,\overline{\mathbf{X}_S}$, $\overline{\mathbf{X}_S^{\mathbf{T}}}\mathbf{X}_S$, and $\mathbf{X}_S^{\mathbf{T}}\overline{\mathbf{X}_S}$ are all equal to $|S|\overline{\mathbf{x}_S^{\mathbf{T}}}\,\overline{\mathbf{x}_S}$. That is, $\mathbf{X}_S^{\mathbf{T}}\overline{\mathbf{X}_S} = \overline{\mathbf{X}_S^{\mathbf{T}}}\mathbf{X}_S = \overline{\mathbf{X}_S^{\mathbf{T}}}\,\overline{\mathbf{X}_S} = |S|\overline{\mathbf{x}_S^{\mathbf{T}}}\,\overline{\mathbf{x}_S}$.*

*Proof.* As proof of Lemma 9 considering $\mathbf{X}_S$ in place of $\mathbf{Y}_S$. □

**Proposition 7.** *Let $S \subseteq R$ and $\mathbf{X}_S^{\mathbf{c}}$ and $\mathbf{Y}_S^{\mathbf{c}}$ be centerings per Definition 3. Then $\mathbf{X}_S^{\mathbf{cT}}\mathbf{Y}_S^{\mathbf{c}} = \mathbf{X}_S^{\mathbf{T}}\mathbf{Y}_S - |S|\overline{\mathbf{x}_S^{\mathbf{T}}}\,\overline{\mathbf{y}_S}$.*

*Proof.* We expand according to Definition 3, using that transposition is distributive, and from Lemma 9 use that $\mathbf{X}_S^{\mathbf{T}}\overline{\mathbf{Y}_S} = \overline{\mathbf{X}_S^{\mathbf{T}}}\mathbf{Y}_S = \overline{\mathbf{X}_S^{\mathbf{T}}}\,\overline{\mathbf{Y}_S} = |S|\overline{\mathbf{x}_S^{\mathbf{T}}}\,\overline{\mathbf{y}_S}$ to derive

$$\begin{aligned}
\mathbf{X}_S^{\mathbf{cT}}\mathbf{Y}_S^{\mathbf{c}} &= (\mathbf{X}_S - \overline{\mathbf{X}_S})^{\mathbf{T}}(\mathbf{Y}_S - \overline{\mathbf{Y}_S}) \\
&= \mathbf{X}_S^{\mathbf{T}}\mathbf{Y}_S - \mathbf{X}_S^{\mathbf{T}}\overline{\mathbf{Y}_S} - \overline{\mathbf{X}_S^{\mathbf{T}}}\mathbf{Y}_S + \overline{\mathbf{X}_S^{\mathbf{T}}}\,\overline{\mathbf{Y}_S} \\
&= \mathbf{X}_S^{\mathbf{T}}\mathbf{Y}_S - |S|\overline{\mathbf{x}_S^{\mathbf{T}}}\,\overline{\mathbf{y}_S}.
\end{aligned}$$

□

**Proposition 8.** *Let $S \subseteq R$ and $\mathbf{X}_S^{\mathbf{c}}$ be a centering per Definition 3. Then $\mathbf{X}_S^{\mathbf{cT}}\mathbf{X}_S^{\mathbf{c}} = \mathbf{X}_S^{\mathbf{T}}\mathbf{X}_S - |S|\overline{\mathbf{x}_S^{\mathbf{T}}}\,\overline{\mathbf{x}_S}$.*

*Proof.* As proof of Proposition 7 using Lemma 10 instead of Lemma 9. □

As mentioned above, Lindgren et al. [16] describe a method for centering validation partition submatrices to support such preprocessing efficiently during cross-validation. The following result uses our notation to express their Equation (20) and is used to show that their method does not satisfy Definition 3 and may lead to leakage by preprocessing.

**Lemma 11.** *The method by Lindgren et al. [16] for obtaining a centered matrix product for training partition $T$ is given by $(\mathbf{X}_T^{\mathbf{T}}\mathbf{X}_T)_{ij} - N\overline{\mathbf{x}}_i\overline{\mathbf{x}}_j(2 - 1/P) + |T|\overline{\mathbf{x}}_i\overline{\mathbf{x}_{Tj}} + |T|\overline{\mathbf{x}}_j\overline{\mathbf{x}_{Ti}}$.*

*Proof.* Let $V$ be a validation partition, $T = R \setminus V$ a training partition, and $(i,j)$ an entry in the matrix product. Equation (20) of Lindgren et al. [16] states the centered matrix product for $T$ is

$$(\mathbf{X}^{\mathbf{T}}\mathbf{X})_{ij} - (\mathbf{X}_V^{\mathbf{T}}\mathbf{X}_V)_{ij} - \overline{\mathbf{x}}_i\sum_{n\in V}\mathbf{X}_{nj} - \overline{\mathbf{x}}_j\sum_{n\in V}\mathbf{X}_{ni} + N/P\cdot\overline{\mathbf{x}}_i\overline{\mathbf{x}}_j.$$

Using Lemma 2 (the fractionated subparts insight due to Lindgren et al. [16]), we can collapse the two left-most terms, and since $V = R \setminus T$, we can express the sums over $V$ using $R$ and $T$. Then, following the definition of mean row vectors and factoring, we arrive at the conclusion,

$$\begin{aligned}
&(\mathbf{X}_T^{\mathbf{T}}\mathbf{X}_T)_{ij} - \overline{\mathbf{x}}_i\sum_{n\in V}\mathbf{X}_{nj} - \overline{\mathbf{x}}_j\sum_{n\in V}\mathbf{X}_{ni} + N/P\cdot\overline{\mathbf{x}}_i\overline{\mathbf{x}}_j \\
&= (\mathbf{X}_T^{\mathbf{T}}\mathbf{X}_T)_{ij} - \overline{\mathbf{x}}_i\left(\sum_{n\in R}\mathbf{X}_{nj} - \sum_{n\in T}\mathbf{X}_{nj}\right) \\
&\qquad - \overline{\mathbf{x}}_j\left(\sum_{n\in R}\mathbf{X}_{ni} - \sum_{n\in T}\mathbf{X}_{ni}\right) + N/P\cdot\overline{\mathbf{x}}_i\overline{\mathbf{x}}_j \\
&= (\mathbf{X}_T^{\mathbf{T}}\mathbf{X}_T)_{ij} - \overline{\mathbf{x}}_iN\overline{\mathbf{x}}_j + \overline{\mathbf{x}}_i|T|\overline{\mathbf{x}_{Tj}} - \overline{\mathbf{x}}_jN\overline{\mathbf{x}}_i + \overline{\mathbf{x}}_j|T|\overline{\mathbf{x}_{Ti}} + N/P\cdot\overline{\mathbf{x}}_i\overline{\mathbf{x}}_j \\
&= (\mathbf{X}_T^{\mathbf{T}}\mathbf{X}_T)_{ij} - N\overline{\mathbf{x}}_i\overline{\mathbf{x}}_j - N\overline{\mathbf{x}}_j\overline{\mathbf{x}}_i + N/P\cdot\overline{\mathbf{x}}_i\overline{\mathbf{x}}_j + |T|\overline{\mathbf{x}}_i\overline{\mathbf{x}_{Tj}} + |T|\overline{\mathbf{x}}_j\overline{\mathbf{x}_{Ti}} \\
&= (\mathbf{X}_T^{\mathbf{T}}\mathbf{X}_T)_{ij} - N\overline{\mathbf{x}}_i\overline{\mathbf{x}}_j(2 - 1/P) + |T|\overline{\mathbf{x}}_i\overline{\mathbf{x}_{Tj}} + |T|\overline{\mathbf{x}}_j\overline{\mathbf{x}_{Ti}}.
\end{aligned}$$

□

Under our definition of centering, we can apply Proposition 8 which states that an entry $(i,j)$ in the centered matrix product for a training partition $T$ is $(\mathbf{X}_T^{\mathbf{T}}\mathbf{X}_T)_{ij} - |T|\left(\overline{\mathbf{x}_T^{\mathbf{T}}}\overline{\mathbf{x}_T}\right)_{ij}$. Together with Lemma 11, we therefore have

$$|T|\left(\overline{\mathbf{x}_T^{\mathbf{T}}}\overline{\mathbf{x}_T}\right)_{ij} = N\overline{\mathbf{x}}_i\overline{\mathbf{x}}_j(2-1/P) - |T|\overline{\mathbf{x}}_i\overline{\mathbf{x}_{Tj}} - |T|\overline{\mathbf{x}}_j\overline{\mathbf{x}_{Ti}}.$$

To derive a contradiction, consider the case where columns $i,j$ in the training partition are centered (mean of 0) but that $i,j$ are not centered in the data set as a whole. This means that $(\overline{\mathbf{x}_T})_i = (\overline{\mathbf{x}_T})_j = \left(\overline{\mathbf{x}_T^{\mathbf{T}}}\overline{\mathbf{x}_T}\right)_{ij} = 0$, $\overline{\mathbf{x}}_i \neq 0$, and $\overline{\mathbf{x}}_j \neq 0$. Therefore, we must have $0 = N\overline{\mathbf{x}}_i\overline{\mathbf{x}}_j(2-1/P)$. This is a contradiction since $\overline{\mathbf{x}}_i \neq 0$, $\overline{\mathbf{x}}_j \neq 0$, and $N \geq P \geq 2$. Moreover, the nonzero quantity subtracted depends on samples not in the training partition, thus leading to information leakage during cross-validation. In contrast, we now show that Algorithm 5 performs centering without such leakage.

**Proposition 9.** *Correctness, Algorithm 5 In step 9 of Algorithm 4,* $\mathrm{cXTX}_T = \mathbf{X}_T^{\mathbf{cT}}\mathbf{X}_T^{\mathbf{c}}$ *and* $\mathrm{cXTY}_T = \mathbf{X}_T^{\mathbf{cT}}\mathbf{Y}_T^{\mathbf{c}}$*, and the variables are identical to the quantities* $\mathrm{cXTX}_T$ *and* $\mathrm{cXTY}_T$ *computed in steps 14 and 15 of Algorithm 5.*

*Proof.* By Lemma 1, both algorithms select validation partition $V$ according to Definition 1. For Algorithm 4, $T = R \setminus V$ by step 5, and step 8 computes $\mathrm{cX}_T$ as $\mathbf{X}_T^{\mathbf{c}} = \mathbf{X}_T - \overline{\mathbf{X}_T}$ and $\mathrm{cY}_T$ as $\mathbf{Y}_S^{\mathbf{c}} = \mathbf{Y}_S - \overline{\mathbf{Y}_S}$ as per Definition 3. Therefore, after step 9, we have that $\mathrm{cXTX}_T$ equals $\mathbf{X}_T^{\mathbf{cT}}\mathbf{X}_T^{\mathbf{c}}$ and $\mathrm{cXTY}_T$ equals $\mathbf{X}_T^{\mathbf{cT}}\mathbf{Y}_T^{\mathbf{c}}$.

Considering now Algorithm 5, we have in step 8 that $\mathrm{XTX}_T = \mathbf{X}_T^{\mathbf{T}}\mathbf{X}_T$ and $\mathrm{XTY}_T = \mathbf{X}_T^{\mathbf{T}}\mathbf{Y}_T$ as shown in Proposition 1. That $\mu\mathrm{x}_T$ in step 11 equals $\overline{\mathbf{x}_T}$ follows from Lemma 7, the selected validation partition $V$, and the quantities computed in steps 9 and 10. Similarly, we have in step 12 that $\mu\mathrm{y}_T$ equals $\overline{\mathbf{y}_T}$ by Lemma 8. Therefore, step 13 computes $\mu\mathrm{xTx}_T$ to equal $|T|\overline{\mathbf{x}_T^{\mathbf{T}}}\overline{\mathbf{x}_T}$ and $\mu\mathrm{xTy}_T$ to equal $|T|\overline{\mathbf{x}_T^{\mathbf{T}}}\overline{\mathbf{y}_T}$. In step 14, it therefore follows from Proposition 8 that $\mathrm{cXTX}_T$ equals $\mathbf{X}_T^{\mathbf{cT}}\mathbf{X}_T^{\mathbf{c}}$, and in step 15, it follows from Proposition 7 that $\mathrm{cXTY}_T$ equals $\mathbf{X}_T^{\mathbf{cT}}\mathbf{Y}_T^{\mathbf{c}}$. Hence, $\mathrm{cXTX}_T$ and $\mathrm{cXTY}_T$ equal the quantities computed in step 9 of Algorithm 4. □

## 4.2 | Computational Complexity

We now analyze the running time and space complexity of Algorithms 4 and 5.

**Proposition 10.** *Algorithm 4 requires* $\Theta(PNK(K+M))$ *operations.*

*Proof.* Storing the inputs and steps 1–7 of Algorithm 2 are equivalent, in terms of running time, to storing the inputs and steps 1–6 and 9 of Algorithm 4, thus requiring $\Theta(PNK(K+M))$ operations as shown in the proof of Proposition 2. Computing the means in step 7 is over $|T_p|$ rows and $K$, respectively, $M$, columns and so requires $\Theta(|T_p|(K+M))$ operations. Step 8 requires $\Theta(|T_p|(K+M))$ operations for matrix subtractions. For $P$ iterations, steps 7 and 8 require $\Theta\left(\sum_{p=1}^{P}|T_p|(K+M)\right)$ operations, which by Lemma 6 is $\Theta(PNK(K+M))$. Thus, Algorithm 4 requires $\Theta(PNK(K+M))$ operations. □

**Proposition 11.** *Algorithm 5 requires* $\Theta(NK(K+M))$ *operations.*

*Proof.* In step 3 computing data set means requires $\Theta(NK) + \Theta(NM) = \Theta(N(K+M))$ operations. The remaining computations in storing the inputs and steps 1–8 are equivalent to storing the inputs and steps 1–8 of Algorithm 3, so as in the proof of Proposition 3 require $\Theta(NK(K+M))$ operations.

Consider a partition $p \in \{1, \dots, P\}$ with validation partition $V_p$ and training partition $T_p$ iterated over in step 4. Computing means in step 9 requires $\Theta(|V_p|(K+M))$ operations, while steps 10–12 require $\Theta(|V_p| + K + M)$ operations to obtain $\overline{\mathbf{x}_T} \in \mathbb{R}^K$ and $\overline{\mathbf{y}_T} \in \mathbb{R}^M$. The matrix multiplications in step 13 require $\Theta(K^2)$ and $\Theta(KM)$ operations, as do the matrix subtractions in steps 14 and 15. Steps 9 through 15 therefore require $\Theta(|V_p|(K+M)) + \Theta(|V_p| + K + M) + \Theta(K^2) + \Theta(KM)$ which is $\Theta(|V_p|(K+M) + K^2 + KM)$. In conducting $P$ iterations and using that $\sum_{p=1}^{P}|V_p| = N$, this gives

$$\begin{aligned}
&\Theta\left(\sum_{p=1}^{P}|V_p|(K+M)+K^2+KM\right)\\
&=\Theta\left((K+M)\sum_{p=1}^{P}|V_p|+\left((K^2+KM)\sum_{p=1}^{P}1\right)\right)\\
&=\Theta\left(N(K+M)+PK^2+PKM\right)
\end{aligned}$$

total operations. Since $P \leq N$ this is at most $O(N(K+M)+NK^2+NKM) = O(N(K+M+K^2+KM)) = O(NK(K+M))$ operations. The total number of operations of Algorithm 5 is therefore $\Theta(NK(K+M)) + O(NK(K+M)) = \Theta(NK(K+M))$. □

**Proposition 12.** *Algorithm 4 requires* $\Theta(P)$ *more operations than Algorithm 5.*

*Proof.* By Propositions 10 and 11, the ratio of operations is $\frac{\Theta(PNK(K+M))}{\Theta(NK(K+M))} = \Theta(P)$. □

With Proposition 12, we have shown that Algorithm 5 is asymptotically faster than Algorithm 4, as also demonstrated for cross-validation without preprocessing in Proposition 4. Below, we show that this comes at no cost in terms of space complexity; that is, Algorithms 4 and 5 are of the same space complexity.

**Proposition 13.** *Algorithm 4 requires storing* $\Theta((K+N)(K+M))$ *entries.*

*Proof.* For step 7, we store the stacked mean row vectors using $|T|K + |T|M$ entries, which is also required for the centerings in step 8. Storing $\mathrm{cXTX}_T$ and $\mathrm{cXTY}_T$ in step 9 in Algorithm 4 requires the same amount of entries as storing $\mathrm{cXTX}_T$ and $\mathrm{cXTY}_T$ in step 7 in Algorithm 2. The remainder of Algorithm 4 is equivalent to Algorithm 2, so as in the proof of Proposition 5 requires storing $\Theta((K+N)(K+M))$ entries. Using $|T| < N$, the total number of entries is $O(N(K+M)) + \Theta((K+N)(K+M)) = \Theta((K+N)(K+M))$. □

**Proposition 14.** *Algorithm 5 requires storing $\Theta((K+N)(K+M))$ entries.*

*Proof.* In step 3, storing $\overline{\mathbf{x}}$ and $\overline{\mathbf{y}}$ requires $\Theta(K+M)$ entries. The remaining storage required for steps 1–8 is equivalent to the storage required by Algorithm 3, so as in the proof of Proposition 6 require $\Theta((N+K)(K+M))$ entries. Step 9 stores $\mu\mathrm{x}_V$ and $\mu\mathrm{y}_V$ requiring $K+M$ entries, the same number of entries needed for $\mu\mathrm{x}_T$ and $\mu\mathrm{y}_T$ in steps 11 and 12. Therefore, steps 9–12 require $\Theta(K+M)$ entries. In step 13, the variables $\mu\mathrm{xTx}_T$ and $\mu\mathrm{xTy}_T$ require $K^2$, respectively, $KM$ entries, as do $\mathrm{cXTX}_T$ and $\mathrm{cXTY}_T$ in steps 14 and 15. Therefore, steps 9–15 store $\Theta(K+M)+2K^2+2KM=\Theta(K(K+M))$ entries, meaning the total required is $\Theta((K+N)(K+M)) + \Theta(K(K+M)) = \Theta((K+N)(K+M))$. $\square$

## 5 | Cross-Validation With Centering and Scaling

We proceed to extend Algorithm 5 to support scaling; that is, we compute $\mathbf{X}_T^{\mathbf{c}\mathbf{T}}\mathbf{X}_T^{\mathbf{c}}$ and $\mathbf{X}_T^{\mathbf{c}\mathbf{T}}\mathbf{Y}_T^{\mathbf{c}}$ where $\mathbf{X}_T^{\mathbf{c}}$ and $\mathbf{Y}_T^{\mathbf{c}}$ are scaled by the column-wise sample standard deviation computed over the training partition $T$. Our method computes the sample standard deviation only once for the entire data set, and then, for each partition $p$, the standard deviation contribution of samples in the validation partition is removed. Adding this important preprocessing extension impacts neither time nor space complexity when compared to Algorithm 5, and to our knowledge, it has not been identified previously.

In what follows, we use Hadamard (element-wise) operators for multiplication ( $\odot$ ), division ( $\oslash$ ), and exponentiation ( $\circ$ ).

**Definition 4.** Let $S \subseteq R$ denote a set of rows such that $|S| \geq 2$. The (Bessels's corrected) sample standard deviation (row) vector $\widehat{\mathbf{x}_S}$ has width $K$ and is the sample standard deviation of each column in $\mathbf{X}_S$, and $\widehat{\mathbf{y}_S}$ has width $M$ and is the sample standard deviation (row) vector of $\mathbf{Y}_S$, given by

$$\widehat{\mathbf{x}_S} = \left( \frac{1}{|S|-1} \sum_{n \in S} \left(\mathbf{X}_n - \overline{\mathbf{x}_S}\right)^{\circ 2} \right)^{\circ \frac{1}{2}} \text{ and}$$
$$\widehat{\mathbf{y}_S} = \left( \frac{1}{|S|-1} \sum_{n \in S} \left(\mathbf{Y}_n - \overline{\mathbf{y}_S}\right)^{\circ 2} \right)^{\circ \frac{1}{2}}.$$

Stacking the sample standard deviation row vectors yields matrices $\widehat{\mathbf{X}_S}$ with dimension $|S| \times K$ and $\widehat{\mathbf{Y}_S}$ with dimension $|S| \times M$ given by

$$\widehat{\mathbf{X}_S} = \begin{bmatrix} \widehat{\mathbf{x}_S} \\ \vdots \\ \widehat{\mathbf{x}_S} \end{bmatrix} \text{ and } \widehat{\mathbf{Y}_S} = \begin{bmatrix} \widehat{\mathbf{y}_S} \\ \vdots \\ \widehat{\mathbf{y}_S} \end{bmatrix}.$$

For brevity, we adopt the convention that $\widehat{\mathbf{X}_S^{\mathbf{T}}}$ and $\widehat{\mathbf{x}_S^{\mathbf{T}}}$ are, respectively, given by $\left(\widehat{\mathbf{X}_S}\right)^{\mathbf{T}}$ and $\left(\overline{\mathbf{x}_S}\right)^{\mathbf{T}}$. We also omit parentheses to denote the $i$'th element of vector $\widehat{\mathbf{x}_S^{\mathbf{T}}}$ by $\widehat{\mathbf{x}_{S_i}^{\mathbf{T}}}$ and the $i$'th element of vector $\widehat{\mathbf{y}_S}$ by $\widehat{\mathbf{y}_{S_i}}$.

**Definition 5.** Let $S \subseteq R$ denote a set of rows such that $|S| \geq 2$ and let $\mathbf{X}_S^{\mathbf{c}}$ and $\mathbf{Y}_S^{\mathbf{c}}$ be given according to Definition 3. The centering and (sample standard deviation) scaling of $\mathbf{X}_S$ is $\mathbf{X}_S^{\mathbf{cs}} = \mathbf{X}_S^{\mathbf{c}} \oslash \widehat{\mathbf{X}_S}$, and the centering and (sample standard deviation) scaling of $\mathbf{Y}_S$ is $\mathbf{Y}_S^{\mathbf{cs}} = \mathbf{Y}_S^{\mathbf{c}} \oslash \widehat{\mathbf{Y}_S}$.

To support centering and scaling of $\mathbf{X}_{T_p}$ and $\mathbf{Y}_{T_p}$ for a partition $p$, we subtract the column-wise means and subsequently divide element-wise by the column-wise sample standard deviations before taking matrix products, namely,

$$\mathbf{X}_{T_p}^{\mathbf{cs}\mathbf{T}}\mathbf{X}_{T_p}^{\mathbf{cs}} = \left(\mathbf{X}_{T_p}^{\mathbf{c}\mathbf{T}} \oslash \widehat{\mathbf{X}_{T_p}^{\mathbf{T}}}\right)\left(\mathbf{X}_{T_p}^{\mathbf{c}} \oslash \widehat{\mathbf{X}_{T_p}}\right) \text{ and}$$
$$\mathbf{X}_{T_p}^{\mathbf{cs}\mathbf{T}}\mathbf{Y}_{T_p}^{\mathbf{cs}} = \left(\mathbf{X}_{T_p}^{\mathbf{c}\mathbf{T}} \oslash \widehat{\mathbf{X}_{T_p}^{\mathbf{T}}}\right)\left(\mathbf{Y}_{T_p}^{\mathbf{c}} \oslash \widehat{\mathbf{Y}_{T_p}}\right).$$

We name these quantities centered and scaled matrix products. The condition $|S| \geq 2$ in the definitions above motivates the following requirement.

**Definition 6.** A partitioning $\mathcal{P}$ of $R$ is *scalable* when $|R \setminus V_p| = |T_p| \geq 2$ for all $p \in \{1, \ldots, P\}$.

In Algorithm 6, we make the immediate extension of Algorithm 4 so that $\mathbf{X}_T^{\mathbf{c}}$ and $\mathbf{Y}_T^{\mathbf{c}}$ are scaled by the sample standard deviations. Then we build on Algorithm 5 in Algorithm 7 with the purpose of efficiently computing $\mathbf{X}_T^{\mathbf{cs}\mathbf{T}}\mathbf{X}_T^{\mathbf{cs}}$ and $\mathbf{X}_T^{\mathbf{cs}\mathbf{T}}\mathbf{Y}_T^{\mathbf{cs}}$ for each partition $p$ with training partition $T = T_p$. In both Algorithms 6 and 7, we consider *scalable* partitionings and

---

**Algorithm 6** Baseline Cross-Validation Algorithm with Centering and Scaling

---

**Input:** $\mathbf{X} \in \mathbb{R}^{N \times K}$, $\mathbf{Y} \in \mathbb{R}^{N \times M}$, $\mathcal{P}$ is a scalable partitioning of $R = \{1, \ldots, N\}$

1: $R \leftarrow \{1, \ldots, N\}$
2: Let $\mathcal{V}$ be the result of executing Algorithm 1 with $\mathcal{P}$ and $R$ as input
3: **for** each partition $p \in \{1, \ldots, P\}$ **do**
4: $\quad V \leftarrow \mathcal{V}[p]$
5: $\quad T \leftarrow R \setminus V$
6: $\quad \mathrm{X}_T \leftarrow \mathbf{X}_T$, $\mathrm{Y}_T \leftarrow \mathbf{Y}_T$
7: $\quad \mu\mathrm{X}_T \leftarrow \overline{\mathbf{X}_T}$, $\mu\mathrm{Y}_T \leftarrow \overline{\mathbf{Y}_T}$
8: $\quad \mathrm{cX}_T \leftarrow \mathrm{X}_T - \mu\mathrm{X}_T$, $\mathrm{cY}_T \leftarrow \mathrm{Y}_T - \mu\mathrm{Y}_T$ $\qquad \triangleright$ Obtain $\mathbf{X}_T^{\mathbf{c}}$ and $\mathbf{Y}_T^{\mathbf{c}}$.
9: $\quad \sigma\mathrm{X}_T \leftarrow \widehat{\mathbf{X}_T}$, $\sigma\mathrm{Y}_T \leftarrow \widehat{\mathbf{Y}_T}$
10: $\quad$ Replace with 1 any entry in $\sigma\mathrm{X}_T$ and $\sigma\mathrm{Y}_T$ that is 0.
11: $\quad \mathrm{csX}_T \leftarrow \mathrm{cX}_T \oslash \sigma\mathrm{X}_T$, $\mathrm{csY}_T \leftarrow \mathrm{cY}_T \oslash \sigma\mathrm{Y}_T$ $\qquad \triangleright$ Obtain $\mathbf{X}_T^{\mathbf{cs}}$ and $\mathbf{Y}_T^{\mathbf{cs}}$
12: $\quad \mathrm{csXTX}_T \leftarrow \mathrm{csX}_T^{\mathbf{T}}\mathrm{csX}_T$, $\mathrm{csXTY}_T \leftarrow \mathrm{csX}_T^{\mathbf{T}}\mathrm{csY}_T$ $\qquad \triangleright$ Obtain $\mathbf{X}_T^{\mathbf{cs}\mathbf{T}}\mathbf{X}_T^{\mathbf{cs}}$ and $\mathbf{X}_T^{\mathbf{cs}\mathbf{T}}\mathbf{Y}_T^{\mathbf{cs}}$
13: **end for**

---

**Algorithm 7** Fast Cross-Validation Algorithm with Centering and Scaling

**Input:** $\mathbf{X} \in \mathbb{R}^{N \times K}$, $\mathbf{Y} \in \mathbb{R}^{N \times M}$, $\mathcal{P}$ is a scalable partitioning of $R = \{1, \dots, N\}$

1: $R \leftarrow \{1, \dots, N\}$
2: Let $\mathcal{V}$ be the result of executing Algorithm 1 with $\mathcal{P}$ and $R$ as input
3: $\mathrm{XTX} \leftarrow \mathbf{X}^{\mathbf{T}}\mathbf{X}$, $\mathrm{XTY} \leftarrow \mathbf{X}^{\mathbf{T}}\mathbf{Y}$, $\mu\mathrm{x} \leftarrow \overline{\mathbf{x}}$, $\mu\mathrm{y} \leftarrow \overline{\mathbf{y}}$
4: $\Sigma\mathrm{X} \leftarrow \sum_{n \in R} \mathbf{X}_n$, $\Sigma\mathrm{Y} \leftarrow \sum_{n \in R} \mathbf{Y}_n$, $\Sigma sq\mathrm{X} \leftarrow \sum_{n \in R} \left(\mathbf{X}_n^{\circ 2}\right)$, $\Sigma sq\mathrm{Y} \leftarrow \sum_{n \in R} \left(\mathbf{Y}_n^{\circ 2}\right)$
5: **for** each partition $p \in \{1, \dots, P\}$ **do**
6: $V \leftarrow \mathcal{V}[p]$
7: $\mathrm{X}_V \leftarrow \mathbf{X}_V$, $\mathrm{Y}_V \leftarrow \mathbf{Y}_V$
8: $\mathrm{XTX}_V \leftarrow \mathrm{X}_V^{\mathbf{T}}\mathrm{X}_V$, $\mathrm{XTY}_V \leftarrow \mathrm{X}_V^{\mathbf{T}}\mathrm{Y}_V$
9: $\mathrm{XTX}_T \leftarrow \mathrm{XTX} - \mathrm{XTX}_V$, $\mathrm{XTY}_T \leftarrow \mathrm{XTY} - \mathrm{XTY}_V$ ▷ Obtain $\mathbf{X}_T^{\mathbf{T}}\mathbf{X}_T$ and $\mathbf{X}_T^{\mathbf{T}}\mathbf{Y}_T$
10: $\mu\mathrm{x}_V \leftarrow \overline{\mathbf{x}_V}$, $\mu\mathrm{y}_V \leftarrow \overline{\mathbf{y}_V}$
11: $N_V \leftarrow |V|$, $N_T \leftarrow N - N_V$
12: $\mu\mathrm{x}_T \leftarrow \frac{N}{N_T}\mu\mathrm{x} - \frac{N_V}{N_T}\mu\mathrm{x}_V$ ▷ Obtain $\overline{\mathbf{x}_T}$
13: $\mu\mathrm{y}_T \leftarrow \frac{N}{N_T}\mu\mathrm{y} - \frac{N_V}{N_T}\mu\mathrm{y}_V$ ▷ Obtain $\overline{\mathbf{y}_T}$
14: $\mu\mathrm{xTx}_T \leftarrow N_T \left(\mu\mathrm{x}_T^{\mathbf{T}}\mu\mathrm{x}_T\right)$, $\mu\mathrm{xTy}_T \leftarrow N_T \left(\mu\mathrm{x}_T^{\mathbf{T}}\mu\mathrm{y}_T\right)$ ▷ Obtain $|T|\overline{\mathbf{x}_T}^{\mathbf{T}}\overline{\mathbf{x}_T}$ and $|T|\overline{\mathbf{x}_T}^{\mathbf{T}}\overline{\mathbf{y}_T}$
15: $\mathrm{cXTX}_T \leftarrow \mathrm{XTX}_T - \mu\mathrm{xTx}_T$ ▷ Obtain $\mathbf{X}_T^{\mathbf{c}\mathbf{T}}\mathbf{X}_T^{\mathbf{c}}$
16: $\mathrm{cXTY}_T \leftarrow \mathrm{XTY}_T - \mu\mathrm{xTy}_T$ ▷ Obtain $\mathbf{X}_T^{\mathbf{c}\mathbf{T}}\mathbf{Y}_T^{\mathbf{c}}$
17: $\Sigma\mathrm{X}_V \leftarrow \sum_{n \in V} \mathrm{X}_n$, $\Sigma\mathrm{Y}_V \leftarrow \sum_{n \in V} \mathrm{Y}_n$
18: $\Sigma\mathrm{X}_T \leftarrow \Sigma\mathrm{X} - \Sigma\mathrm{X}_V$, $\Sigma\mathrm{Y}_T \leftarrow \Sigma\mathrm{Y} - \Sigma\mathrm{Y}_V$ ▷ Obtain $\Sigma_{n \in T}\mathbf{X}_n$ and $\Sigma_{n \in T}\mathbf{Y}_n$
19: $\Sigma sq\mathrm{X}_V \leftarrow \sum_{n \in V} \left(\mathbf{X}_n^{\circ 2}\right)$, $\Sigma sq\mathrm{Y}_V \leftarrow \sum_{n \in V} \left(\mathbf{Y}_n^{\circ 2}\right)$
20: $\Sigma sq\mathrm{X}_T \leftarrow \Sigma sq\mathrm{X} - \Sigma sq\mathrm{X}_V$ ▷ Obtain $\Sigma_{n \in T}\left(\mathbf{X}_n^{\circ 2}\right)$
21: $\Sigma sq\mathrm{Y}_T \leftarrow \Sigma sq\mathrm{Y} - \Sigma sq\mathrm{Y}_V$ ▷ Obtain $\Sigma_{n \in T}\left(\mathbf{Y}_n^{\circ 2}\right)$
22: $\sigma\mathrm{x}_T \leftarrow \left(\frac{1}{N_T - 1}\left(N_T \mu\mathrm{x}_T^{\circ 2} + \Sigma sq\mathrm{X}_T - 2\mu\mathrm{x}_T \odot \Sigma\mathrm{X}_T\right)\right)^{\circ \frac{1}{2}}$ ▷ Obtain $\widehat{\mathbf{x}_T}$
23: $\sigma\mathrm{y}_T \leftarrow \left(\frac{1}{N_T - 1}\left(N_T \mu\mathrm{y}_T^{\circ 2} + \Sigma sq\mathrm{Y}_T - 2\mu\mathrm{y}_T \odot \Sigma\mathrm{Y}_T\right)\right)^{\circ \frac{1}{2}}$ ▷ Obtain $\widehat{\mathbf{y}_T}$
24: Replace with 1 any entry in $\sigma\mathrm{x}_T$ and $\sigma\mathrm{y}_T$ that is 0.
25: $\sigma\mathrm{xTx}_T \leftarrow \sigma\mathrm{x}_T^{\mathbf{T}}\sigma\mathrm{x}_T$, $\sigma\mathrm{xTy}_T \leftarrow \sigma\mathrm{x}_T^{\mathbf{T}}\sigma\mathrm{y}_T$ ▷ Obtain $\widehat{\mathbf{x}_T^{\mathbf{T}}}\widehat{\mathbf{x}_T}$ and $\widehat{\mathbf{x}_T^{\mathbf{T}}}\widehat{\mathbf{y}_T}$
26: $\mathrm{csXTX}_T \leftarrow \mathrm{cXTX}_T \oslash \sigma\mathrm{xTx}_T$ ▷ Obtain $\mathbf{X}_T^{\mathbf{cs}\mathbf{T}}\mathbf{X}_T^{\mathbf{cs}}$
27: $\mathrm{csXTY}_T \leftarrow \mathrm{cXTY}_T \oslash \sigma\mathrm{xTy}_T$ ▷ Obtain $\mathbf{X}_T^{\mathbf{cs}\mathbf{T}}\mathbf{Y}_T^{\mathbf{cs}}$
28: **end for**

employ the standard practice of replacing 0-entries with 1-entries when using the standard deviation vectors and matrices for scaling (to avoid division by zero). Correctness and complexity results follow.

## 5.1 | Correctness

As in Section 4, enabling fast sample standard deviation-scaling requires us to compute this quantity exclusively over rows in the validation partition. In Lemmas 12 and 13, we show how the training partition sample standard deviations $\widehat{\mathbf{x}_T^{\mathbf{T}}}$ and $\widehat{\mathbf{y}_T}$ are rearrangeable into simpler constituent terms. While these quantities are based on the training partition, Lemma 14 shows they are computable by removing the contribution from validation partition samples from the constituent terms over the entire set of samples, which we need only compute once.

**Lemma 12.** *Let $S \subseteq R$, $|S| \geq 2$, and $\widehat{\mathbf{x}_S}$ be the sample standard deviation vector. Then*

$$\widehat{\mathbf{x}_S} = \left(\frac{1}{|S|-1}\left(|S|\overline{\mathbf{x}_S}^{\circ 2} + \sum_{n \in S}\left(\mathbf{X}_n^{\circ 2}\right) - 2\overline{\mathbf{x}_S} \odot \sum_{n \in S}\mathbf{X}_n\right)\right)^{\circ \frac{1}{2}}.$$

*Proof.* Expanding the Hadamard exponentiation in $\widehat{\mathbf{x}_S}$ from Definition 4 yields

$$\left(\frac{1}{|S|-1}\sum_{n \in S}\left(\left(\mathbf{X}_n - \overline{\mathbf{x}_S}\right)^{\circ 2}\right)\right)^{\circ \frac{1}{2}}$$
$$= \left(\frac{1}{|S|-1}\sum_{n \in S}\left(\mathbf{X}_n^{\circ 2} + \overline{\mathbf{x}_S}^{\circ 2} - 2\overline{\mathbf{x}_S} \odot \mathbf{X}_n\right)\right)^{\circ \frac{1}{2}}.$$

By commutativity of addition, the distributivity of Hadamard multiplication over addition, simplifying the summation, and rearranging terms, we get

$$\left(\frac{1}{|S|-1}\left(\sum_{n \in S}\left(\overline{\mathbf{x}_S}^{\circ 2}\right) + \sum_{n \in S}\left(\mathbf{X}_n^{\circ 2}\right) - \sum_{n \in S} 2\overline{\mathbf{x}_S} \odot \mathbf{X}_n\right)\right)^{\circ \frac{1}{2}}$$
$$= \left(\frac{1}{|S|-1}\left(|S|\overline{\mathbf{x}_S}^{\circ 2} + \sum_{n \in S}\left(\mathbf{X}_n^{\circ 2}\right) - 2\overline{\mathbf{x}_S} \odot \sum_{n \in S}\mathbf{X}_n\right)\right)^{\circ \frac{1}{2}}.$$

□

**Lemma 13.** *Let $S \subseteq R$, $|S| \geq 2$, and $\widehat{\mathbf{y}_S}$ be the sample standard deviation vector. Then*

$$\widehat{\mathbf{y}_S} = \left(\frac{1}{|S|-1}\left(|S|\overline{\mathbf{y}_S}^{\circ 2} + \sum_{n \in S}\left(\mathbf{Y}_n^{\circ 2}\right) - 2\overline{\mathbf{y}_S} \odot \sum_{n \in S}\mathbf{Y}_n\right)\right)^{\circ \frac{1}{2}}.$$

*Proof.* As proof of Lemma 12, considering $\widehat{\mathbf{y}_S}$ in place of $\widehat{\mathbf{x}_S}$. □

**Lemma 14.** *Let $p$ be a partition, $T = T_p$ a training partition, and $V = V_p$ a validation partition. Then*

$$
\begin{aligned}
\Sigma_{n\in T}\mathbf{X}_n &= \Sigma_{n\in R}\mathbf{X}_n - \Sigma_{n\in V}\mathbf{X}_n,\\
\Sigma_{n\in T}\mathbf{Y}_n &= \Sigma_{n\in R}\mathbf{Y}_n - \Sigma_{n\in V}\mathbf{Y}_n,\\
\Sigma_{n\in T}\left(\mathbf{X}_n^{\circ 2}\right) &= \Sigma_{n\in R}\left(\mathbf{X}_n^{\circ 2}\right) - \Sigma_{n\in V}\left(\mathbf{X}_n^{\circ 2}\right), \text{ and}\\
\Sigma_{n\in T}\left(\mathbf{Y}_n^{\circ 2}\right) &= \Sigma_{n\in R}\left(\mathbf{Y}_n^{\circ 2}\right) - \Sigma_{n\in V}\left(\mathbf{Y}_n^{\circ 2}\right).
\end{aligned}
$$

*Proof.* By Definition 1, we have that $T \cap V = \emptyset$ and $T \cup V = R$. Therefore,

$$
\sum_{n\in T}\mathbf{X}_n + \sum_{n\in V}\mathbf{X}_n = \sum_{n\in T\cup V}\mathbf{X}_n = \sum_{n\in R}\mathbf{X}_n;
$$

hence, $\Sigma_{n\in T}\mathbf{X}_n = \Sigma_{n\in R}\mathbf{X}_n - \Sigma_{n\in V}\mathbf{X}_n$. The remaining equalities are shown similarly. □

With Propositions 15 and 16, we can efficiently compute $\mathbf{X}_S^{\mathbf{csT}}\mathbf{X}_S^{\mathbf{cs}}$ and $\mathbf{X}_S^{\mathbf{csT}}\mathbf{Y}_S^{\mathbf{cs}}$ given the (already) centered $\mathbf{X}_S^{\mathbf{cT}}\mathbf{X}_S^{\mathbf{c}}$ and $\mathbf{X}_S^{\mathbf{cT}}\mathbf{Y}_S^{\mathbf{c}}$, by using only an outer vector product and Hadamard division over the matrix products.

**Proposition 15.** *Let $S \subseteq R$ and $|S| \geq 2$, then $\mathbf{X}_S^{\mathbf{csT}}\mathbf{Y}_S^{\mathbf{cs}} = \left(\mathbf{X}_S^{\mathbf{cT}}\mathbf{Y}_S^{\mathbf{c}}\right) \oslash \left(\widehat{\mathbf{x}_S^{\mathbf{T}}}\widehat{\mathbf{y}_S}\right)$.*

*Proof.* Let $(i,j)$ be an entry in $\mathbf{X}_S^{\mathbf{csT}}\mathbf{Y}_S^{\mathbf{cs}}$. By Definition 4, each *column* in $\widehat{\mathbf{X}_S^{\mathbf{T}}}$ is the column vector $\widehat{\mathbf{x}_S^{\mathbf{T}}}$; thus, for all $n \in S$, we have $\widehat{\mathbf{x}_S^{\mathbf{T}}} = \widehat{\mathbf{X}_S^{\mathbf{T}}}_{*n}$ and so $\widehat{\mathbf{x}_S^{\mathbf{T}}}_i = \widehat{\mathbf{X}_S^{\mathbf{T}}}_{in}$. Similarly, for all $n \in S$, each *row* in $\widehat{\mathbf{Y}_S}$ is the row vector $\widehat{\mathbf{y}_S} = \widehat{\mathbf{Y}_S}_{n*}$ and so $\widehat{\mathbf{y}_S}_j = \widehat{\mathbf{Y}_S}_{nj}$. Using distributivity of division over summation, we get

$$
\begin{aligned}
\left(\mathbf{X}_S^{\mathbf{csT}}\mathbf{Y}_S^{\mathbf{cs}}\right)_{ij} &= \left(\left(\mathbf{X}_S^{\mathbf{cT}} \oslash \widehat{\mathbf{X}_S^{\mathbf{T}}}\right)\left(\mathbf{Y}_S^{\mathbf{c}} \oslash \widehat{\mathbf{Y}_S}\right)\right)_{ij}\\
&= \sum_{n\in S}\frac{\mathbf{X}_{S_{in}}^{\mathbf{cT}}}{\widehat{\mathbf{X}_S^{\mathbf{T}}}_{in}}\frac{\mathbf{Y}_{S_{nj}}^{\mathbf{c}}}{\widehat{\mathbf{Y}_S}_{nj}} = \sum_{n\in S}\frac{\mathbf{X}_{S_{in}}^{\mathbf{cT}}\mathbf{Y}_{S_{nj}}^{\mathbf{c}}}{\widehat{\mathbf{x}_S^{\mathbf{T}}}_i\widehat{\mathbf{y}_S}_j}\\
&= \frac{\sum_{n\in S}\mathbf{X}_{S_{in}}^{\mathbf{cT}}\mathbf{Y}_{S_{nj}}^{\mathbf{c}}}{\widehat{\mathbf{x}_S^{\mathbf{T}}}_i\widehat{\mathbf{y}_S}_j} = \frac{\left(\mathbf{X}_S^{\mathbf{cT}}\mathbf{Y}_S^{\mathbf{c}}\right)_{ij}}{\left(\widehat{\mathbf{x}_S^{\mathbf{T}}}\widehat{\mathbf{y}_S}\right)_{ij}}\\
&= \left(\left(\mathbf{X}_S^{\mathbf{cT}}\mathbf{Y}_S^{\mathbf{c}}\right) \oslash \left(\widehat{\mathbf{x}_S^{\mathbf{T}}}\widehat{\mathbf{y}_S}\right)\right)_{ij},
\end{aligned}
$$

showing the equality. □

**Proposition 16.** *Let $S \subseteq R$ and $|S| \geq 2$, then $\mathbf{X}_S^{\mathbf{csT}}\mathbf{X}_S^{\mathbf{cs}} = \left(\mathbf{X}_S^{\mathbf{cT}}\mathbf{X}_S^{\mathbf{c}}\right) \oslash \left(\widehat{\mathbf{x}_S^{\mathbf{T}}}\widehat{\mathbf{x}_S}\right)$.*

*Proof.* As proof of Proposition 16 considering $\mathbf{X}_S$ in place of $\mathbf{Y}_S$. □

While our targets in this section are $\mathbf{X}_S^{\mathbf{csT}}\mathbf{X}_S^{\mathbf{cs}}$ and $\mathbf{X}_S^{\mathbf{csT}}\mathbf{Y}_S^{\mathbf{cs}}$ (particularly $S = T_p$), we note that these results generalize to the case where $\mathbf{X}_S$ and $\mathbf{Y}_S$ are not centered. Exclusively preprocessing $\mathbf{X}_S$ and $\mathbf{Y}_S$ using sample standard deviation-scaling is therefore possible, as is preprocessing only one of $\mathbf{X}_S$ or $\mathbf{Y}_S$. We revisit these variations in Section 6.

**Proposition 17.** *Correctness of Algorithm 7 In step 12 of Algorithm 6, $\text{csXTX}_T = \mathbf{X}_T^{\mathbf{csT}}\mathbf{X}_T^{\mathbf{cs}}$ and $\text{csXTY}_T = \mathbf{X}_T^{\mathbf{csT}}\mathbf{Y}_T^{\mathbf{cs}}$, and the variables are identical to the quantities $\text{XTX}_T$ and $\text{XTY}_T$ computed in steps 26 and 27 of Algorithm 7.*

*Proof.* Due to Proposition 1, both algorithms select $T = T_p$ and $V = V_p$ as per Definition 1 (a scalable partitioning is a valid partitioning). Since Algorithm 6 is equivalent to Algorithm 4 up to and including step 8, we have as in Proposition 9 that $\text{cX}_T$ equals $\mathbf{X}_T^{\mathbf{c}}$ and $\text{cY}_T$ equals $\mathbf{Y}_T^{\mathbf{c}}$. Steps 9–11 find the sample standard deviation matrices (replacing 0-entries with 1-entries) and apply them to the centered matrices so that $\text{csX}_T$ is $\mathbf{X}_T^{\mathbf{cs}} = \mathbf{X}_T^{\mathbf{c}} \oslash \widehat{\mathbf{X}_T}$ and $\text{csY}_T$ is $\mathbf{Y}_T^{\mathbf{cs}} = \mathbf{Y}_T^{\mathbf{c}} \oslash \widehat{\mathbf{Y}_T}$ as per Definition 5. Therefore, after step 12, we have that $\text{csXTX}_T$ equals $\mathbf{X}_T^{\mathbf{csT}}\mathbf{X}_T^{\mathbf{cs}}$ and $\text{csXTY}_T$ equals $\mathbf{X}_T^{\mathbf{csT}}\mathbf{Y}_T^{\mathbf{cs}}$.

Consider now Algorithm 7. In steps 15 and 16, we have that $\text{cXTX}_T$ equals $\mathbf{X}_T^{\mathbf{cT}}\mathbf{X}_T^{\mathbf{c}}$ and $\text{cXTY}_T$ equals $\mathbf{X}_T^{\mathbf{cT}}\mathbf{Y}_T^{\mathbf{c}}$ as shown in Proposition 9. That $\Sigma\text{X}_T$ equals $\sum_{n\in T}\mathbf{X}_n$ and $\Sigma\text{Y}_T$ equals $\sum_{n\in T}\mathbf{Y}_n$ as computed in step 18 follows from Lemma 14 given the quantities computed in step 17. Also, from Lemma 14 follows that $\Sigma sq\text{X}_T$ equals $\sum_{n\in T}\left(\mathbf{X}_n^{\circ 2}\right)$ and $\Sigma sq\text{Y}_T$ equals $\sum_{n\in T}\left(\mathbf{Y}_n^{\circ 2}\right)$ as computed in steps 20 and 21 due to the quantities computed in step 19. By Lemma 12 and the quantities computed in steps 12, 18, and 20, it follows that after step 22, $\sigma\text{x}_T$ equals $\widehat{\mathbf{x}_T}$. Similarly, applying Lemma 13 and the quantities computed in steps 13, 18, and 21, we have that after step 23, $\sigma\text{y}_T$ equals $\widehat{\mathbf{y}_T}$.

Replacing 0-entries of $\sigma\text{x}_T$ and $\sigma\text{y}_T$ in step 24 means that these vectors are the rows in matrices $\sigma\text{X}_T$ and $\sigma\text{Y}_T$ after step in 10 of Algorithm 6 according to Definition 4. Therefore, after step 25, $\sigma\text{xTx}_T$ and $\sigma\text{xTy}_T$ are equal to $\widehat{\mathbf{x}_T^{\mathbf{T}}}\widehat{\mathbf{x}_T}$ and $\widehat{\mathbf{x}_T^{\mathbf{T}}}\widehat{\mathbf{y}_T}$, respectively. Thus, for step 26, we can apply Proposition 16 to see that $\text{csXTX}_T$ equals $\mathbf{X}_T^{\mathbf{csT}}\mathbf{X}_T^{\mathbf{cs}}$, and Proposition 15 for step 27 to see that $\text{csXTY}_T$ equals $\mathbf{X}_T^{\mathbf{csT}}\mathbf{Y}_T^{\mathbf{cs}}$. □

## 5.2 | Computational Complexity

We now analyze the running time and space complexities of Algorithms 6 and 7. Our results show that centering and scaling can be enabled for cross-validation with no asymptotic impact on time and space complexity compared to cross-validation without preprocessing.

**Proposition 18.** *Algorithm 6 requires $\Theta(PNK(K+M))$ operations.*

*Proof.* Steps 1 through 8 of Algorithm 6 are equivalent to steps 1 through 8 of Algorithm 4. Moreover, step 12 of Algorithm 6 requires the same amount of operations as step 9 of Algorithm 4. Operations required to store the inputs are also equivalent between Algorithm 4 and Algorithm 6. Therefore, as in Proposition 10, these steps require $\Theta(PNK(K+M))$ operations.

For the cost of the remaining steps 9–11, consider any partition $p$ iterated over in step 3. To compute the sample standard deviations row vectors $\widehat{\mathbf{x}_{T_p}}$ and $\widehat{\mathbf{y}_{T_p}}$ in step 9, we sum over $|T_p|$ rows and $K$, respectively $M$, columns. We perform $2K$, respectively $2M$, operations to subtract and square for each row. We take the Hadamard square root of the resulting vectors and then multiply the scaling factor requiring $2K$, respectively $2M$, operations. Computing $1/(|T_p|-1))$ requires two operations. Stacking the resulting vectors to construct $\widehat{\mathbf{Y}_{T_p}}$ and $\widehat{\mathbf{Y}_{T_p}}$ requires $|T_p|(K+M)$ operations. In total, step 9 requires $2+5|T_p|(K+M)$ operations. Step 10 requires $|T_p|(K+M)$ operations to locate 0-entries in $\widehat{\mathbf{X}_{T_p}}$ and $\widehat{\mathbf{Y}_{T_p}}$ and at most $|T_p|(K+M)$ operations to replace them with 1-entries. Step 11 performs Hadamard division requiring $|T_p|(K+M)$ operations. The total for steps 9–11 is $2+7|T_p|(K+M)+O(N+M)=\Theta\big(|T_p|(K+M)\big)$ operations.

Iterating over $P$ partitions means steps 9–11 require $\Theta\Big(\sum_{p=1}^{P}|T_p|(K+M)\Big)$ operations, which by Lemma 6 simplifies to $\Theta(PN(K+M))$ operations. The total number of operations required by Algorithm 6 is therefore $\Theta(PN(K+M))+\Theta(PNK(K+M)) =\Theta(PNK(K+M))$. □

**Proposition 19.** *Algorithm 7 requires $\Theta(NK(K+M))$ operations.*

*Proof.* In step 4, computing $\Sigma_{n\in R}\mathbf{X}_n$, $\Sigma_{n\in R}\mathbf{Y}_n$, $\Sigma_{n\in R}\big(\mathbf{X}_n^{\circ 2}\big)$ and $\Sigma_{n\in R}\big(\mathbf{Y}_n^{\circ 2}\big)$ require $3N(K+M)=\Theta(N(K+M))$ operations. The remaining computations that store inputs and perform steps 1 through 16 are equivalent to storing inputs and steps 1–15 in Algorithm 5, so as in Proposition 11 require $\Theta(NK(K+M))$ operations. Steps 1–16 in Algorithm 7 therefore require $\Theta(N(K+M))+\Theta(NK(K+M))=\Theta(NK(K+M))$ operations.

For the remaining steps, consider any partition $p$. Step 17 requires $|V_p|(K+M)$ operations. Step 18 requires $K+M$ operations. Step 19 requires $2|V_p|(K+M)$ operations. Steps 20 and 21 require $K+M$ operations. Steps 22 and 23 require $6(K+M)+2$ operations. Step 24 requires $K+M$ operations to locate 0-entries in $\sigma \mathrm{x}_T$ and $\sigma \mathrm{y}_T$ and at most $K+M$ operations to replace them with 1-entries. The matrix products in step 25 require $\Theta(K(K+M))$ operations. Steps 26 and 27 require $K(K+M)$ operations. The total for steps 17–27 is $3|V_p|(K+M)+9(K+M)+2+O(K+M)+\Theta(K(K+M))+K(K+M)$ which is $\Theta(|V_p|(K+M)+K(K+M))$ operations. Iterating over $P$ partitions and using that $\sum_{p=1}^{P}|V_p|=N$, steps 17–27 require

$$\begin{aligned}
&\Theta\left(\sum_{p=1}^{P}|V_p|(K+M)+K(K+M))\right)\\
&=\Theta\left((K+M)\sum_{p=1}^{P}|V_p|+K\right)\\
&=\Theta((K+M)(N+PK))\\
&=\Theta(N(K+M))+\Theta(PK(K+M))
\end{aligned}$$

operations. Since $P\leq N$, this is $O(NK(K+M))$ operations. Thus, Algorithm 7 requires a total of $\Theta(NK(K+M))+O(NK(K+M))=\Theta(NK(K+M))$ operations. □

**Proposition 20.** *Algorithm 6 requires $\Theta(P)$ more operations than Algorithm 7.*

*Proof.* By Propositions 18 and 19, the ratio of operations is $\frac{\Theta(PNK(K+M))}{\Theta(NK(K+M))}=\Theta(P)$. □

With Proposition 20, we have shown that Algorithm 7 is asymptotically faster (shaving a factor of $\Theta(P)$) than the baseline Algorithm 6. This corresponds with the results against the baselines without preprocessing or with centering only, as shown in Propositions 4 and 12. Similarly, we can show this incurs no cost in terms of space complexity; that is, Algorithms 6 and 7 are of same space complexity.

**Proposition 21.** *Algorithm 6 requires storing $\Theta((K+N)(K+M))$ entries.*

*Proof.* Step 9 requires $|T|(K+M)$ entries to store $\sigma \mathrm{X}_T$ and $\sigma \mathrm{Y}_T$. Step 10 requires no additional entries to modify $\sigma \mathrm{X}_T$ and $\sigma \mathrm{Y}_T$ in-place. Step 11 requires $|T|(K+M)$ to store $\mathrm{csX}_T$ and $\mathrm{csY}_T$. Storing $\mathrm{csXTX}_T$ and $\mathrm{csXTY}_T$ in step 12 in Algorithm 6 requires the same amount of entries as storing $\mathrm{cXTX}_T$ and $\mathrm{cXTY}_T$ in step 9 in Algorithm 4. The remainder of Algorithm 6 is equivalent to Algorithm 4, so as in the proof of Proposition 13 requires storing $\Theta((K+N)(K+M))$ entries. Summing each contribution and using $|T|<N$, the total number of matrix entries is $2|T|(K+M)+\Theta((K+N)(K+M))=\Theta((K+N)(K+M))$. □

**Proposition 22.** *Algorithm 7 requires storing $\Theta((K+N)(K+M))$ entries.*

*Proof.* In steps 3 and 4, storing $\sum_{n\in R}\mathbf{X}_n$, $\sum_{n\in R}\mathbf{Y}_n$, $\sum_{n\in R}\big(\mathbf{X}_n^{\circ 2}\big)$ and $\sum_{n\in R}\big(\mathbf{Y}_n^{\circ 2}\big)$ requires $2(K+M)=\Theta(K+M)$ entries. The remaining storage required for steps 1–16 is equivalent to the storage required by Algorithm 5, so as in the proof of Proposition 14 require $\Theta((N+K)(K+M))$ entries. Steps 17–23 require $5(K+M)$ entries to store $\Sigma \mathrm{X}_V$, $\Sigma \mathrm{Y}_V$, $\Sigma \mathrm{X}_T$, $\Sigma \mathrm{Y}_T$, $\Sigma sq\mathrm{X}_V$, $\Sigma sq\mathrm{Y}_V$, $\Sigma sq\mathrm{X}_T$, $\Sigma sq\mathrm{Y}_T$, $\sigma \mathrm{x}_T$, and $\sigma \mathrm{y}_T$. Step 24 requires no additional entries. Steps 25–27 require $2K(K+M)$ entries to store $\sigma \mathrm{xTx}_T$, $\sigma \mathrm{xTy}_T$, $\mathrm{csXTX}_T$, and $\mathrm{csXTY}_T$. The total number of entries is therefore $\Theta(K+M)+\Theta((N+K)(K+M))+5(K+M)+2K(K+M)$ which is $\Theta((K+N)(K+M))$. □

We remark that for all three of our fast algorithms, we have proved time bounds that match computing $\mathbf{X}^{\mathbf{T}}\mathbf{X}$ and $\mathbf{X}^{\mathbf{T}}\mathbf{Y}$ only and space bounds that match storing $\mathbf{X}$, $\mathbf{Y}$, $\mathbf{X}^{\mathbf{T}}\mathbf{X}$, and $\mathbf{X}^{\mathbf{T}}\mathbf{Y}$. In other words, $P$-fold cross-validation can be performed in the same time and space (asymptotically) as it takes to compute just one fold, which is supported up to a practical constant by our benchmarks in Section 7.

## 6 | Preprocessing Combinations

In previous sections, we considered algorithms that constitute three combinations of preprocessing, namely, none, centering both $\mathbf{X}_T$ and $\mathbf{Y}_T$, and centering and scaling both $\mathbf{X}_T$ and $\mathbf{Y}_T$. Our fast methods also work if we want to only center, only scale, or center and scale, either $\mathbf{X}_T$ or $\mathbf{Y}_T$. Note that when both centering and scaling, our propositions rely on centering first.

In the case of $\mathbf{X}_T^{\mathbf{T}}\mathbf{X}_T$, there are four different preprocessing combinations, each resulting in a distinct matrix product. This contrasts the case for $\mathbf{X}_T^{\mathbf{T}}\mathbf{Y}_T$ where the $2^4=16$ different preprocessing combinations result in only eight distinct matrix

**TABLE 1** | The effects on $\mathbf{X}_T^{\mathbf{T}}\mathbf{Y}_T$ of applying all possible combinations of centering and scaling on $\mathbf{X}_T$ and $\mathbf{Y}_T$.

<table>
<tr><th></th><th></th><th colspan="3">Centering</th></tr>
<tr><th>Scaling</th><th>No centering</th><th>Center $\mathbf{X}_T$</th><th>Center $\mathbf{Y}_T$</th><th>Center both</th></tr>
<tr><td>No scaling</td><td>$\mathbf{X}_T^{\mathbf{T}}\mathbf{Y}_T$</td><td colspan="3">$\mathbf{X}_T^{\mathbf{T}}\mathbf{Y}_T - |T|\left(\overline{\mathbf{x}_T^{\mathbf{T}}}\,\overline{\mathbf{y}_T}\right)$</td></tr>
<tr><td>Scale $\mathbf{X}_T$</td><td>$\left(\mathbf{X}_T^{\mathbf{T}}\mathbf{Y}_T\right) \oslash \left(\widehat{\mathbf{x}_T^{\mathbf{T}}}\mathbf{1}_M\right)$</td><td colspan="3">$\left(\mathbf{X}_T^{\mathbf{T}}\mathbf{Y}_T - |T|\left(\overline{\mathbf{x}_T^{\mathbf{T}}}\,\overline{\mathbf{y}_T}\right)\right) \oslash \left(\widehat{\mathbf{x}_T^{\mathbf{T}}}\mathbf{1}_M\right)$</td></tr>
<tr><td>Scale $\mathbf{Y}_T$</td><td>$\left(\mathbf{X}_T^{\mathbf{T}}\mathbf{Y}_T\right) \oslash \left(\mathbf{1}_K^{\mathbf{T}}\widehat{\mathbf{y}_T}\right)$</td><td colspan="3">$\left(\mathbf{X}_T^{\mathbf{T}}\mathbf{Y}_T - |T|\left(\overline{\mathbf{x}_T^{\mathbf{T}}}\,\overline{\mathbf{y}_T}\right)\right) \oslash \left(\mathbf{1}_K^{\mathbf{T}}\widehat{\mathbf{y}_T}\right)$</td></tr>
<tr><td>Scale both</td><td>$\left(\mathbf{X}_T^{\mathbf{T}}\mathbf{Y}_T\right) \oslash \left(\widehat{\mathbf{x}_T^{\mathbf{T}}}\,\widehat{\mathbf{y}_T}\right)$</td><td colspan="3">$\left(\mathbf{X}_T^{\mathbf{T}}\mathbf{Y}_T - |T|\left(\overline{\mathbf{x}_T^{\mathbf{T}}}\,\overline{\mathbf{y}_T}\right)\right) \oslash \left(\widehat{\mathbf{x}_T^{\mathbf{T}}}\,\widehat{\mathbf{y}_T}\right)$</td></tr>
</table>

*Note:* Note that centering $\mathbf{X}_T$, $\mathbf{Y}_T$, or both leads to the same effect on $\mathbf{X}_T^{\mathbf{T}}\mathbf{Y}_T$.

products. This follows from Proposition 9, since as long as either $\mathbf{X}_T$ or $\mathbf{Y}_T$ is centered, the matrix product $\mathbf{X}_T^{\mathbf{T}}\mathbf{Y}_T$ is centered. Conversely, this is not the case for scaling. We show the 8 distinct matrix products $\mathbf{X}_T^{\mathbf{T}}\mathbf{Y}_T$ for all 16 preprocessing combinations in Table 1, where $\mathbf{1}_L$ denotes a row vector of length $L$ with all entries being 1 and where (preprocessed) products are represented using Propositions 9 and 15.

Considering $\mathbf{X}_T^{\mathbf{T}}\mathbf{X}_T$ and $\mathbf{X}_T^{\mathbf{T}}\mathbf{Y}_T$ simultaneously, and if we insist preprocessing of $\mathbf{X}_T$ is the same for both matrix products, there are 16 different preprocessing combinations. In this case, 12 are distinct since centering only $\mathbf{Y}_T$ results in $\mathbf{X}_T^{\mathbf{T}}\mathbf{Y}_T$ being centered while $\mathbf{X}_T^{\mathbf{T}}\mathbf{X}_T$ is not. These insights are relevant to situations where centering is applied asymmetrically in model building.

## 7 | Benchmarks

In Sections 3–5, we have proven the asymptotic efficiency of Algorithms 3, 5, 7 compared to their baseline alternatives. We now demonstrate their practical efficiency as well. Benchmarks are based on the implementation by Engstrøm [21] (Apache 2.0 license), where the bulk of scientific computations take place using NumPy [47].

Before diving into results for execution time, we note some implementation details by Engstrøm [21] that describe the concrete realization of the pseudo-code. For a training partition $T$, centering using Algorithms 4 and 6 (baselines) is realized by storing the mean row vectors $\overline{\mathbf{x}_T}$ and $\overline{\mathbf{y}_T}$ rather than the matrices $\overline{\mathbf{X}_T}$ and $\overline{\mathbf{Y}_T}$. Similarly, for scaling in Algorithm 6, we store sample standard deviation vectors $\widehat{\mathbf{x}_T}$ and $\widehat{\mathbf{x}_T}$ rather than $\widehat{\mathbf{X}_T}$ and $\widehat{\mathbf{Y}_T}$. When applying mean centering, the fast algorithms, Algorithm 5 (step 13) and Algorithm 7 (step 14) obtain $|T|\overline{\mathbf{x}_T^{\mathbf{T}}}\,\overline{\mathbf{y}_T}$ and $|T|\overline{\mathbf{x}_T^{\mathbf{T}}}\,\overline{\mathbf{x}_T}$ (using Propositions 7 and 8) by computing the outer vector product before multiplying by $N_T = |T|$. This is typically the more numerically stable option. However, it requires $|T|(K + M)$ multiplications, whereas first multiplying by $|T|$ can be done with $\min(K, M)$ multiplications (either choice does not impact time complexity analysis).

Our benchmarks are over data set matrices $\mathbf{X}$ and $\mathbf{Y}$ with $N = 100{,}000 = 10^5$, $K = 500$, and $M = 10$. Matrices are randomly initialized with 64-bit values and are identical for all runs. We compute matrix products over $P \in \{3, 5, 10, 100, 1000, 10000, 100000\}$ partitions and use a balanced partitioning so that validation partitions are of equal size (except when $P = 3$ where one validation partition contains one sample more than the others). All benchmarks use the same hardware (AMD Ryzen 9 5950X, 3.4GHz), and we restrict NumPy to a single CPU.

Figure 1 compares the execution time of the baseline and the fast algorithms (so without preprocessing, with centering, with centering and scaling) for each value of $P$. The execution time of baseline algorithms grows linearly with $P$, supporting our results for their time complexity. For Algorithm 3 and $P \geq 10^4$, as well as Algorithms 5 and 7 for $P \geq 10^3$, the results indicate a dependence on $P$ in terms of execution time, which is not supported by our time complexity results.

We profile the implementation and determine operations whose relative cost increases as $P$ grows, finding those to be computing $\mathrm{XTX}_V$ (when $P \geq 10^4$), and the outer vector products $\mu\mathrm{x}_T^{\mathbf{T}}\mu\mathrm{x}_T$ and $\sigma\mathrm{x}_T^{\mathbf{T}}\sigma\mathrm{x}_T$ (when $P \geq 10^3$), indicating the relative cost primarily increases for operations involving $\mathbf{X}_V^{\mathbf{T}}\mathbf{X}_V$. Since $P$ directly impacts $|V|$ and $K = 500$, we measure the execution time of NumPy when computing the matrix product $\mathbf{M}^{\mathbf{T}}\mathbf{M}$ where $\mathbf{M} \in \mathbb{R}^{|V| \times 500}$. We normalize execution times by $|V|$ and divide by the smallest normalized execution time we observe. This gives standardized costs $c$ that describe the execution time of matrix multiplication as a function of $|V|$, where $c$ close to 1 means practical constants are stable.

The standardized costs are summarized in Table 2 and illustrate that the overhead of computing $\mathbf{M}^{\mathbf{T}}\mathbf{M}$ grows as $|V|$ decreases and this in a manner consistent with our benchmarks of the fast algorithms. For the fast algorithms when $P = 10^4$ ($P = 10^5$), $\mathrm{XTX}_V$ requires matrix multiplications that in practice are about 2.5 (17) times more costly than when $P = 10^3$ ($P = 10^4$). With centering and scaling, we compute $2P$ outer vector products[2] corresponding to $|V| = 1$, which Table 2 shows has a large practical constant for our choice of $K$. This accounts for the observable dependency on $P \geq 10^4$ in the execution time of Algorithm 3 and dependency on $P \geq 10^3$ for the execution times of Algorithms 5 and 7.

Notwithstanding the practical overhead of computing many relatively small matrix products, our fast algorithms dominate the execution time of baseline algorithms for all values of $P$ and by more than two orders of magnitude when $P \geq 10^3$. For leave-one-out cross-validation with centering and scaling specifically, execution time drops from close to a day to around a minute in this benchmark.

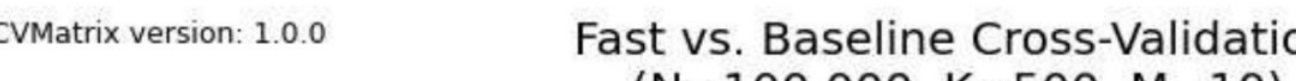

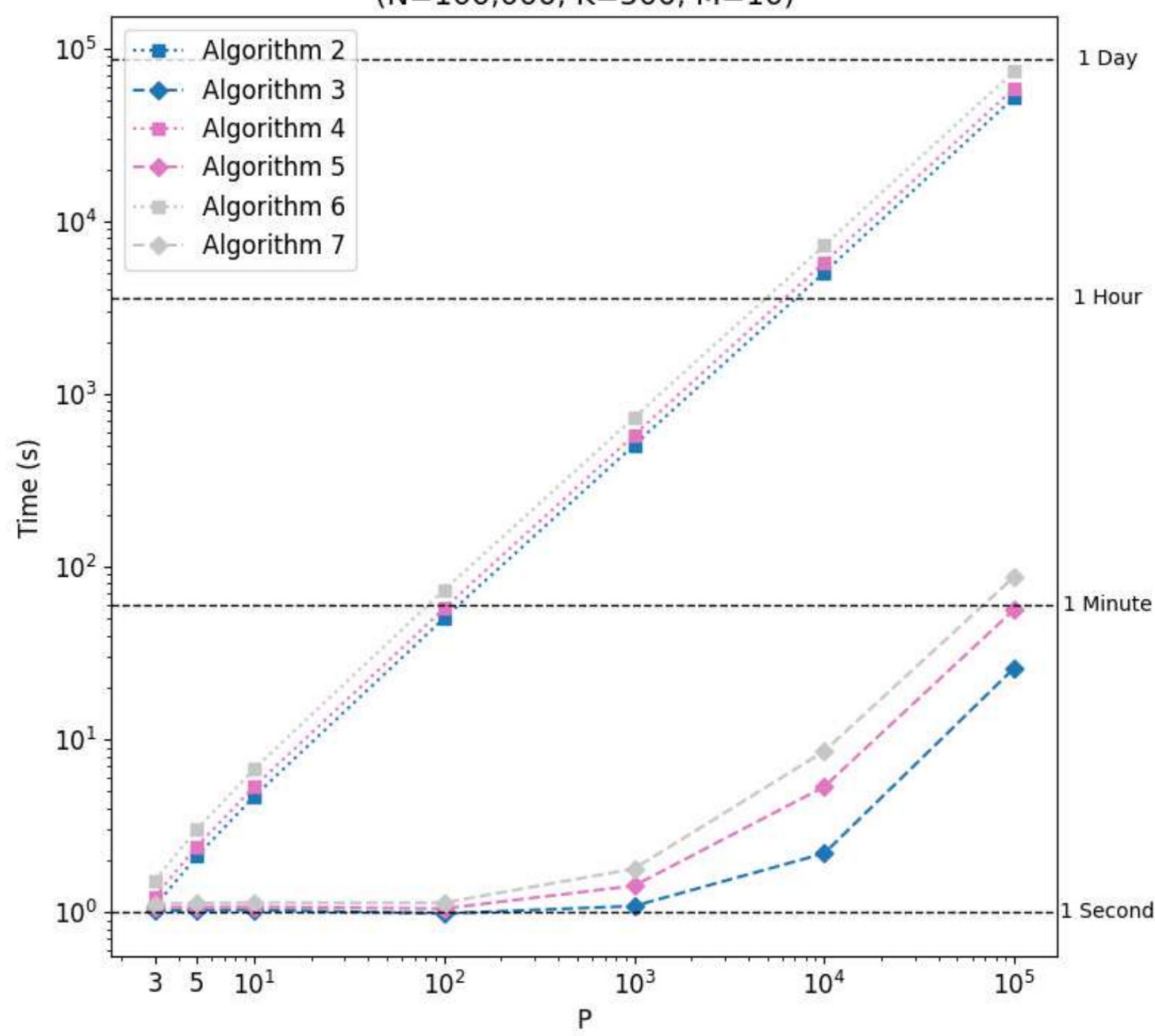

**FIGURE 1** | A log–log plot of execution time as a function of $P$ for computing $\mathbf{X}^{\mathbf{T}}\mathbf{X}$, $\mathbf{X}^{\mathbf{T}}\mathbf{Y}$, and their centered and scaled variants, with the baseline and fast algorithms as implemented in `cvmatrix` [21]. Fast algorithms dominate the baselines by a factor of $\Theta(P)$. The execution time dependence on larger $P$ for the fast algorithms is caused by practical constants from computing many relatively small matrix products.

**TABLE 2** | Standardized cost $c$ of computing $\mathbf{M}^{\mathbf{T}}\mathbf{M}$ with $\mathbf{M} \in \mathbb{R}^{|V| \times 500}$, where $P$ corresponds to the number of partitions yielding $|V|$ when $N = 100{,}000$.

| $P$ | 20 | 100 | 200 | 1000 | 2000 | 10,000 | 20,000 | 50,000 | 100,000 |
|---|---|---|---|---|---|---|---|---|---|
| $\lvert V \rvert$ | 5000 | 1000 | 500 | 100 | 50 | 10 | 5 | 2 | 1 |
| $c \leq$ | 1 | 1.01 | 1.02 | 1.2 | 1.4 | 3 | 5 | 11 | 52 |

*Note:* Shaded columns correspond to a configuration of $P$ in Figure 1.

## 8 | Conclusion

In this paper, we introduced algorithms that significantly accelerate, in theory and practice, partition-based cross-validation for machine learning models. This has application for methods such as PCA, PCR, RR, OLS, and PLS, particularly on tall data sets. Under the fold-based partitioning scheme, we show correctness of our fast algorithms (Propositions 9 and 17), which is a novel contribution since, as we show in Section 4.1, fast alternatives in the literature induce leakage of mean and sample standard deviation statistics between training and validation partitions.

The algorithms we developed work by computing $\mathbf{X}^{\mathbf{T}}\mathbf{X}$, $\mathbf{X}^{\mathbf{T}}\mathbf{Y}$, means, and sample standard deviations only once for the entire data set in the beginning and then, for each cross-validation fold, remove the contribution of the validation partition. This approach eliminates redundant computations existing in the overlap between training partitions and is the same shortcut employed by Lindgren et al. [16], but now correctly generalized for centering and scaling. A key result is Lemma 9 which ultimately collapses certain combinations of center preprocessing and states that $\overline{\mathbf{X}_S}^{\mathbf{T}}\overline{\mathbf{Y}_S}$ is the same as both $\overline{\mathbf{X}_S}^{\mathbf{T}}\mathbf{Y}_S$ and $\mathbf{X}_S^{\mathbf{T}}\overline{\mathbf{Y}_S}$ for any $S \subseteq R$, meaning either can be subtracted from $\mathbf{X}_S^{\mathbf{T}}\mathbf{Y}_S$ to perform centering. This insight applies beyond the fold-based partitioning scheme.

Time and space are shown to be independent of the number of folds (Proposition 20) in the most involved case with both centering and scaling and so has the same time complexity as computing $\mathbf{X}^{\mathbf{T}}\mathbf{X}$ and $\mathbf{X}^{\mathbf{T}}\mathbf{Y}$ (Proposition 19) and space complexity equivalent to storing $\mathbf{X}$, $\mathbf{Y}$, $\mathbf{X}^{\mathbf{T}}\mathbf{X}$, and $\mathbf{X}^{\mathbf{T}}\mathbf{Y}$ (Proposition 22). Our benchmarks of execution time (using implementation by Engstrøm [21], Apache 2.0 license) validate our time complexity results. Therefore, our methods enable practically efficient $P$-fold cross-validation with proper centering and scaling, potentially decreasing execution time by orders of magnitude, making them a valuable tool for robust model selection.

### Peer Review

The peer review history for this article is available at https://www.webofscience.com/api/gateway/wos/peer-review/10.1002/cem.70008.

### Endnotes

[1] The product $\mathbf{X}^{\mathbf{T}}\mathbf{X}$ is sometimes referred to as the covariance matrix [1], $\mathbf{X}^{\mathbf{T}}\mathbf{Y}$ the cross-covariance matrix [2], and both as kernel matrices [3]. Formally, the sample covariance $\mathbf{X}^{\mathbf{T}}\mathbf{X}$ and sample cross-covariance $\mathbf{X}^{\mathbf{T}}\mathbf{Y}$ matrices are computed over centered $\mathbf{X}$ and $\mathbf{Y}$ and then scaled by $\frac{1}{N-1}$. The Pearson correlation matrix also uses scaling (z-standardizing) of $\mathbf{X}$ and $\mathbf{Y}$. We are concerned with different combinations of centering and scaling and, as such, refer to $\mathbf{X}^{\mathbf{T}}\mathbf{X}$ and $\mathbf{X}^{\mathbf{T}}\mathbf{Y}$ simply as matrix products.

[2] Computing $\mathbf{M}^{\mathbf{T}}\mathbf{M}$ is faster, in absolute terms, when $\mathbf{M} \in \mathbb{R}^{2\times500}$ than when $\mathbf{M} \in \mathbb{R}^{1\times500}$. This insight could be readily applied to roughly halve the practical constants associated with centering and scaling and improve the case of leave-one-out cross-validation. We observe for $\mathbf{M} \in \mathbb{R}^{1\times500}$ in our setup that NumPy v1.26.4 no longer uses optimized multiply-accumulate instructions but rather exclusively optimized multiply instructions.